\documentclass{article}

\usepackage{microtype}
\usepackage{graphicx}
\usepackage{subcaption}
\usepackage{booktabs} 

\usepackage{hyperref}

\usepackage[preprint]{icml2026}

\usepackage{amsmath}
\usepackage{amssymb}
\usepackage{mathtools}
\usepackage{amsthm}

\usepackage{enumitem}
\setlist{nosep,leftmargin=*}

\usepackage{fvextra}
\DefineVerbatimEnvironment{WrappedVerbatim}{Verbatim}{breaklines=true,fontsize=\small}

\usepackage[capitalize,noabbrev]{cleveref}

\theoremstyle{plain}

\theoremstyle{definition}

\theoremstyle{remark}

\usepackage[textsize=tiny]{todonotes}

\icmltitlerunning{Test-Time Scaling of Divergent--Convergent Reasoning}

\begin{document}

\twocolumn[
  \icmltitle{Test-Time Scaling of Divergent--Convergent Reasoning}



  \icmlsetsymbol{equal}{*}

  \begin{icmlauthorlist}
    \icmlauthor{Bo Wen}{equal,Enkira}
    \icmlauthor{Yuhao Chen}{equal,queens}
    \icmlauthor{Erhan Bilal}{Enkira}
    \icmlauthor{Carla Agurto Rios}{Enkira}
    \icmlauthor{Chen Wang}{ibm}
    \icmlauthor{Junchen Jiang}{uchicago,tensormesh}
  \end{icmlauthorlist}

  \icmlaffiliation{queens}{School of Computing, Queen's University, Kingston, Ontario, Canada}
  \icmlaffiliation{Enkira}{Enkira.ai, USA}
  \icmlaffiliation{ibm}{IBM T.J. Watson Research Center, USA}
  \icmlaffiliation{uchicago}{Department of Computer Science, University of Chicago, USA}
  \icmlaffiliation{tensormesh}{Tensormesh Inc., USA}

  \icmlcorrespondingauthor{Bo Wen}{bwen@hogarthian.com}

  \icmlkeywords{Machine Learning, Test-Time Compute Scaling, Multi-Agent Reasoning}

  \vskip 0.3in
]



\printAffiliationsAndNotice{\icmlEqualContribution}

\begin{abstract}

Test-time compute can substantially improve Large Language Model (LLM) reasoning performance, yet how and when additional compute helps remains poorly understood. We study \emph{Divergent--Convergent Reasoning} (DCR), a simple two-phase primitive consisting of an exploration phase that generates multiple candidate solutions followed by a convergent reconciliation phase. We present three core results. First, we show that even a single reconciliation step can reliably amplify \emph{correct minority reports}: across datasets, DCR often recovers the correct answer when correct exploration outputs are in the minority, a regime where majority voting fails. Second, we introduce \emph{recursive DCR}, an autoregressive reconciliation system that iteratively analyzes disagreements and allocates additional test-time compute. Recursive DCR achieves higher accuracy than fixed-compute baselines—reaching \textbf{93.3\%} on AIME 2024 and \textbf{92.0\%} on AIME 2025—while using roughly \textbf{27\% less compute} on average, demonstrating that attentive resource allocation is superior to uniform scaling. Third, we analyze disagreement among exploration outputs via a simple, training-free dispersion metric. Dispersion reveals a structured relationship between disagreement and test-time gains: in regimes where DCR is effective, higher disagreement among exploration outputs is associated with larger accuracy improvements from reconciliation. Together, these results show that disagreement, often viewed as noise, can be systematically exploited to improve test-time reasoning and reveal emerging scaling laws for agentic LLM systems.
\end{abstract}

\section{Introduction}
Test-time compute can substantially improve LLM reasoning accuracy, often by sampling multiple candidate solutions and reconciling them via aggregation or evaluation. Yet disagreement among solutions is typically treated as noise, and majority voting is often assumed to be optimal. We revisit this assumption and ask: \emph{can disagreement, particularly minority opinions, be systematically exploited to improve reasoning accuracy?} Our approach builds on the intuition that \emph{selection is easier than generation}: recognizing a correct argument among candidates is often a simpler cognitive task than producing it from scratch.

\begin{figure}[ht]
\vskip -0.1in
\begin{center}
\centerline{\includegraphics[width=0.9\columnwidth]{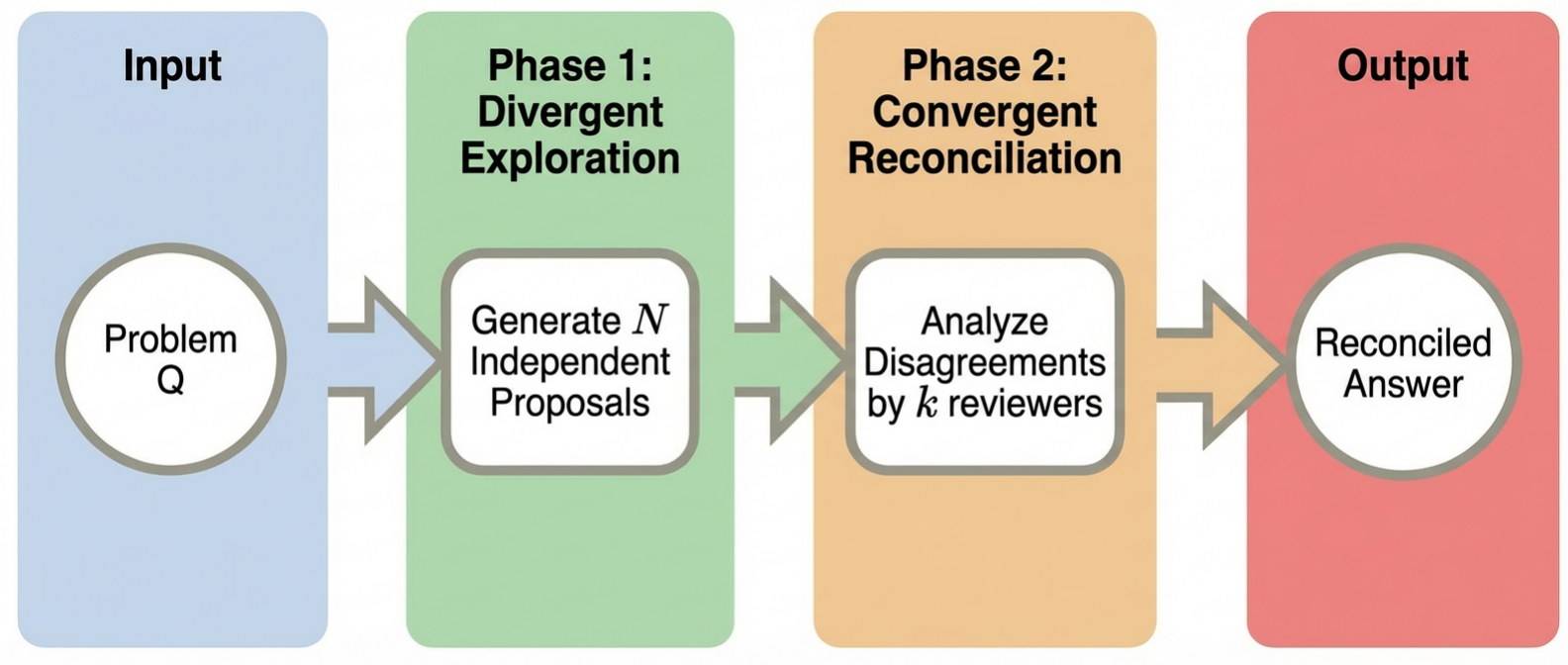}}
\caption{Single-round DCR (linear): generate diverse proposals, then run one reconciliation round with fixed reviewer width \(K\) (default: \(K=25\)).}
\label{fig:dcr_architecture}
\end{center}
\vskip -0.25in
\end{figure}

We study \emph{Divergent--Convergent Reasoning} (DCR), a simple two-phase primitive (Figure~\ref{fig:dcr_architecture}): (i) \textbf{Divergent Exploration} generates diverse independent proposals; (ii) \textbf{Convergent Reconciliation} uses reviewer calls to analyze disagreements and synthesize a single reconciled answer.

\begin{figure*}[ht]
   \vskip -0.1in
   \centering
   \includegraphics[width=0.75\textwidth]{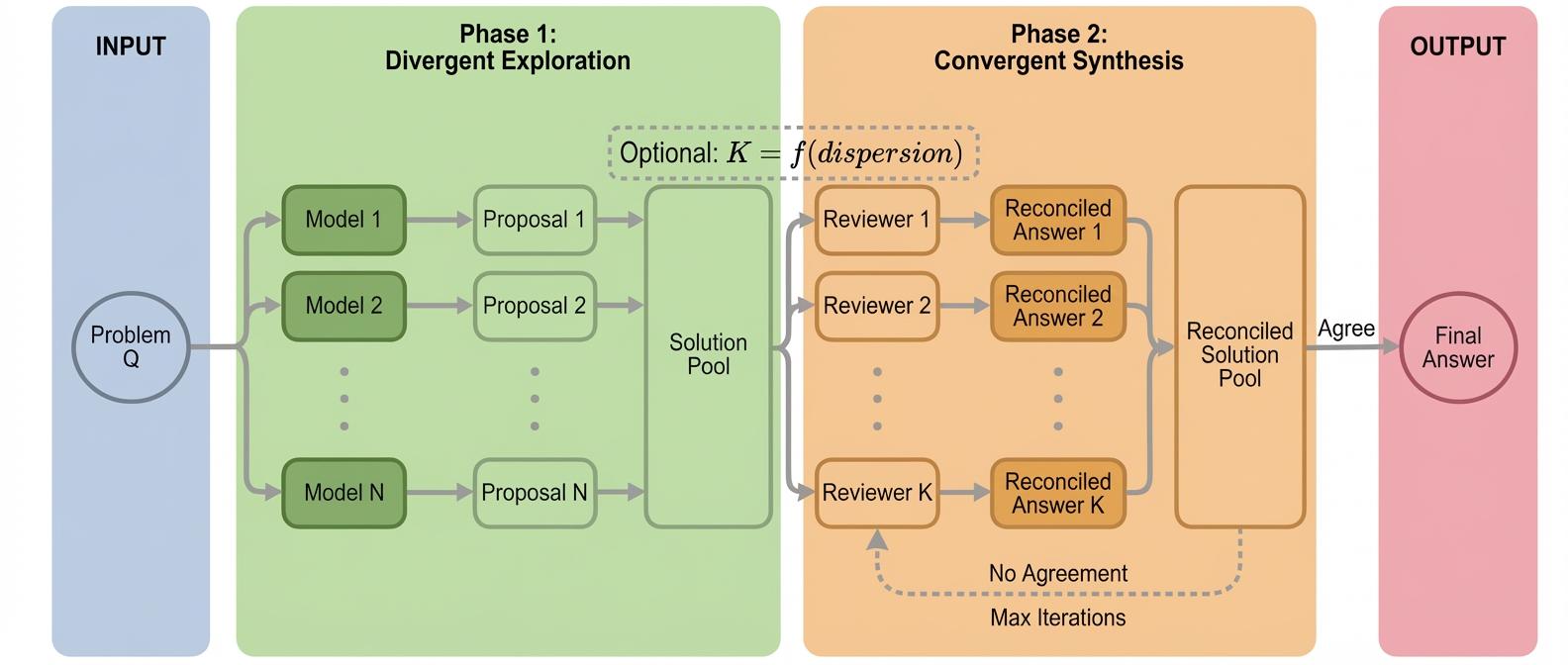}
   \caption{Recursive DCR: starting from an initial proposal pool, the system runs Convergent reconciliation in \emph{rounds} and stops when the $K$ reviewer calls in a round reach \emph{unanimous agreement}, or when a maximum round/call budget is exhausted. In our implementation, the per-round reviewer batch size (reviewer width \(K\)) is fixed (default: \(K=8\)); the \emph{depth} (number of rounds) is therefore task-dependent and self-adjusting. If budget is exhausted without consensus, the system returns a majority answer along with a minority report for downstream review.}
   \label{fig:dcr_recursive_architecture}
   \vskip -0.25in
   \end{figure*}

Our results show that reconciliation can reliably amplify correct minority reports, succeeding precisely where majority vote fails. We further show that recursively analyzing disagreements enables more effective use of additional test-time compute. Finally, the degree of disagreement itself provides a powerful analytical lens for understanding when reconciliation is most beneficial. These findings motivate a closer look at both the \emph{single-round} DCR primitive and its recursive extension, illustrated in Figures~\ref{fig:dcr_architecture} and~\ref{fig:dcr_recursive_architecture}.

\noindent\textbf{Results and Contributions.} (1) \textbf{Minority report matters:} across datasets and settings, \textbf{single-round DCR} (reviewer width \(K=25\)) increases \emph{consistency} over exploration even when correct exploration answers are a minority, showing that reconciliation can amplify correct minority reports where majority voting fails (Table~\ref{tab:result1}). (2) \textbf{Attentive test-time compute via recursion:} we introduce \textbf{recursive DCR}, an autoregressive reconciliation system with unanimous-consent stopping, yielding larger gains than compute-matched aggregation baselines (Figure~\ref{fig:acc_vs_compute}). (3) \textbf{Dispersion as an analytical diagnostic:} a simple training-free dispersion metric over exploration outputs predicts when reconciliation is most effective (Figure~\ref{fig:dispersion_vs_diagnostic}).

We position DCR not as a replacement for verifier-based methods, but as a complementary approach for scenarios where trained verifiers are unavailable or impractical. Such constraints arise naturally in real-world settings such as model routing and budgeted inference, where systems must decide how to spend limited test-time compute without access to immediate ground-truth feedback.

\section{Related Work}

\noindent\textbf{Test-Time Compute Scaling}
Test-time compute scaling has emerged as a central lever for improving LLM reasoning performance. A classical baseline is majority voting and self-consistency \cite{wang2022self}. More broadly, recent work has studied inference scaling laws and compute-efficient inference \cite{snell2024scaling,wu2024inference,sardana2023beyond}, as well as decoding and inference-time algorithms that treat generation as an iterative procedure rather than a single forward pass \cite{welleck2024decoding,chen2024more}. Orthogonally, many systems wrap an LLM with explicit search or refinement procedures, including tree-style deliberate search \cite{yao2023tree} and learning-to-search or refinement paradigms \cite{gandhi2024stream,qu2024recursive}. 

Our work studies a simple and widely used primitive within this space: generating multiple candidate solutions followed by reconciliation. We empirically characterize when additional reconciliation compute is beneficial, rather than proposing new search operators or scaling compute uniformly.

\noindent\textbf{Verifier-Based Methods}
Verifier-based (VB) approaches leverage explicit feedback signals—outcome rewards, verifiers, or learned reward models—to guide search or reinforcement learning \cite{cobbe2021training,uesato2022solving,setlur2024rewarding}. Recent post-training systems based on verification signals have demonstrated strong reasoning performance \cite{deepseekai2025deepseekr1incentivizingreasoningcapability,MoonshotAI,deepscaler2025}. These methods use verification either to shape which reasoning traces models learn to produce or to guide candidate selection at test time.

\citet{setlur2025scaling} formalize an important separation: when reliable verification signals are available for fine-tuning or test-time search, verifier-based methods can make more efficient use of scaled test-time compute than purely verifier-free approaches. Our work is not in conflict with this conclusion. Instead, we study the complementary deployment regime in which a reliable verifier is unavailable prior to answer revelation, yet the system must still allocate a finite inference budget and return a single solution.

\noindent\textbf{Verifier-Free and Multi-Instance Methods}
Verifier-free (VF) approaches rely on supervised distillation, imitation of expert traces, or self-improvement heuristics \cite{muennighoff2025s1,chu2025sft,ye2025limo}. While attractive when verification is expensive or infeasible, such methods inherit challenges arising from heterogeneity in expert reasoning traces and limited control over query-level uncertainty.

Relatedly, multi-agent interaction methods such as debate \cite{du2023improving} exploit structured interaction among model replicas as a source of signal. ReConcile \cite{chen2024reconcile} is a closely related multi-model framework that uses multi-round discussion (with confidence-weighted voting) to reach consensus. DCR differs in emphasis and mechanism: we study a reviewer-style reconciliation primitive (not peer persuasion), analyze \emph{minority-correct amplification} and dispersion, and introduce a simple recursive stopping rule (unanimous-consent) for selective test-time compute.

We provide a detailed paper-by-paper differentiation from closely related verifier-free, early-stopping, and multi-agent aggregation methods in Appendix~\ref{app:prior-art-diff}.

\section{Methodology: Divergent--Convergent Reasoning}

The DCR framework is designed to mimic the process of committee review, separating the generation of diverse ideas from the critical review and reconciliation of those ideas. The process is structured into two main phases, which can be iterated to achieve deeper reasoning.

\noindent\textbf{Problem Setting.} We study inference-time reasoning under a fixed computational budget (measured as total model API calls during exploration and reconciliation) \footnote{The number of API calls is a \emph{controllable} parameter in compute systems, unlike token count, which is often endogenous to the model and task difficulty. This choice also enables clean compute-matched comparisons across aggregation strategies in our report.}, where no verifier or ground-truth feedback is available during inference. Given a task problem $Q$ and stochastic generators producing candidate solutions (proposals) $\{x_n\}_{n=1}^N$, the goal is to return a final answer $\hat{x}$ that maximizes correctness.

\subsection{Phase 1: Divergent Exploration}

Given a problem $Q$, the first phase aims to generate a diverse set of $N$ proposals. In practice, we obtain proposals $\{x_n\}_{n=1}^N$ by sampling one or more LLM replicas independently (possibly a heterogeneous ensemble), using different stochastic seeds/temperatures to encourage diversity. The full set of proposals is collected into a solution pool $S = \{x_n\}_{n=1}^N$. This process is analogous to multiple independent research labs exploring a scientific question from different perspectives.

\subsection{Phase 2: Convergent Reconciliation}

This phase takes the diverse proposals from the solution pool $S$ as input and aims to synthesize a single, high-quality reconciled answer. Instead of a simple majority vote, a new set of reviewer models is tasked with critically analyzing the proposals. In each reconciliation round, we run \(K\) independent reviewer calls (reviewer width \(K\)), indexed by \(k\in\{1,\dots,K\}\). The prompt for these reviewers (see Appendix~\ref{app:prompts}) instructs them to:

\begin{enumerate}
\item \textbf{Identify Consensus Steps:} Steps or reasoning paths that are common across most proposals are likely correct and require only brief verification.
\item \textbf{Pinpoint Divergent Steps:} The exact points where solutions disagree are identified as the most challenging parts of the problem, requiring deep analysis.
\item \textbf{Synthesize a Reconciled Answer (and retain minority reports):} The reviewer must construct a new, coherent solution by combining the verified consensus steps with a newly reasoned path through the divergent steps, while explicitly considering whether a ``minority'' proposal contains a crucial correction that the majority missed.
\end{enumerate}

This process forces the model to engage in metacognition, reasoning about the disagreements themselves to uncover the correct solution path.

\subsection{Iterative Refinement and Adaptive Control}

DCR can be run as a single pass: one Divergent Exploration pool followed by one Convergent Reconciliation round. We also study a \emph{recursive} (multi-round) variant that applies Convergent Reconciliation in \emph{rounds}, feeding each round's reconciled outputs back in as the proposal pool for the next round. This yields an autoregressive control loop that adaptively allocates test-time compute based on whether disagreement persists.

\noindent\textbf{Unanimous-consent stopping.}
In the recursive DCR system studied here, the \textbf{number of reconciliation rounds is not pre-specified}. After each round, we check whether the $K$ reviewer calls in that round have reached \emph{unanimous agreement} (i.e., all $K$ calls produce the same final answer). If unanimity is reached, we \textbf{stop early} and return that answer. If disagreement persists, we run another round, until either (i) unanimity is reached or (ii) a maximum round/call budget is exhausted.

This stopping rule is \emph{verifier-free} and self-adjusting: easy tasks tend to converge quickly, while hard tasks require more rounds (or fail to converge within budget). Beyond the per-round reviewer batch size and the maximum budget, the mechanism requires no additional tuning.

\noindent\textbf{Optional dispersion signals.}
Separately from the core stopping rule, we study \emph{dispersion} of the exploration pool as a training-free, post-hoc diagnostic and as a possible \emph{triage} heuristic (e.g., to set a more conservative max budget, to escalate to a stronger model, or to abstain). We define dispersion formally in Appendix~\ref{app:dispersion} and analyze how it correlates with when reconciliation helps and with overconfident failure modes (Result~3). Dispersion is not required for the core recursive control loop (Section~3; Result~3).

Motivated by the intuition that hard tasks require more deliberation, recursive DCR provides a simple verifier-free mechanism for \emph{selective} test-time compute: easy tasks converge quickly and stop early, while hard tasks naturally consume more rounds (up to a fixed max budget).

\section{Experiments}

We conducted a series of experiments to evaluate the performance of the DCR framework across multiple datasets and models. Our goal was to answer the following questions:

\begin{enumerate}
\item Does Convergent Reconciliation consistently improve upon the initial proposals from Divergent Exploration?
\item How does performance change when mixing proposals from models of varying capabilities?
\item Does the DCR process lead to greater consensus (consistency) among solutions?
\item When does reconciliation provide gains beyond majority vote, especially when correct proposals are a minority or absent? Furthermore, how does recursive reconciliation behave in terms of recovery and convergence?
\end{enumerate}

\subsection{Datasets and Models}

We used four challenging reasoning benchmarks: \textbf{MATH500} \cite{lightman2023let}, \textbf{AIME 2024} \cite{aime2024hf}, \textbf{AIME 2025} \cite{aime2025hf}, and \textbf{MMLU-PRO} \cite{wang2024mmlupro}. For our model ensemble, we selected four LLMs representing a spectrum of capabilities: \textbf{Granite-4-H-Small}, \textbf{Llama-3.3-70B}, \textbf{Llama-4-Maverick-17B-128E}, and \textbf{GPT-OSS-120B} (with only default/medium reasoning efforts, NOT the high effort).

\subsection{Experimental Setup}

We designed three experiments and one compute-matched aggregation baseline:

\begin{enumerate}
\item \textbf{Experiment 1 (Single-model, single-round DCR):} Each model generates a pool of proposals for each problem, and the \emph{same} model performs a single reconciliation round on its \emph{own} pool with reviewer width \(K=25\). We repeat reconciliation for \(T=25\) stochastic trials to estimate reliability under a fixed proposal pool.
\item \textbf{Experiment 2 (Mix-model proposals, single-round DCR):} We simulate collaboration by creating a \emph{mixed proposal pool} derived from all four models. For each problem, we combine the proposals from Experiment 1 and sample a new set of 25 that reflects the average accuracy of the group (as if each model contributed equally). We then run single-round DCR ($K=25$) on this mixed pool using each model as the reviewer. This tests how performance changes when mixing proposals from models of varying capabilities.
\item \textbf{Experiment 3 (Recursive DCR / autoregressive reconciliation):} Starting from a proposal pool, we run multi-round reconciliation until reviewers reach unanimous agreement (early stopping) or a maximum round budget is reached. Recursive DCR can be run with either single-model or mixed proposals; in the main text we report the GPT-OSS-120B setting where proposals are generated by GPT and each reviewer call is also GPT. Our default recursive setting uses reviewer width \textbf{\(K=8\) per round}. This yields a dynamic compute profile: as few as 8 reviewer calls for fast convergence, up to 80 calls for 10 rounds; the ``compute parity'' point with a \(T=25\) single-round baseline is around 3 rounds (\(8\times 3=24\)).
\end{enumerate}

\begin{table*}[t]
\centering
\small
\setlength{\tabcolsep}{4pt}
\renewcommand{\arraystretch}{1.05}
\caption{\textbf{Result 1: Single-round DCR (\(K=25\)) amplifies correct minority reports.} We compare \textbf{Sampling} (exploration baseline) against \textbf{DCR} (reconciliation). \textbf{Trial Acc.} is the average per-attempt accuracy (approximating \textbf{Best-of-N}); \textbf{Consistency} measures reliability (percentage of problems where the method is correct \(>50\%\) of the time, approximating \textbf{Majority Vote}). \textbf{DCR (Single)} reviews the model's own proposals; \textbf{DCR (Mix)} reviews a shared pool of diverse proposals. DCR consistently improves upon Sampling, often turning fragile minority correct answers into stable majority outcomes.}
\label{tab:result1}
\begin{tabular}{llrrr@{\hspace{8pt}}rrr}
\toprule
\textbf{Dataset} & \textbf{Model} & \multicolumn{3}{c}{\textbf{Trial acc. (\%) $\approx$ Best-of-N}} & \multicolumn{3}{c}{\textbf{Consistency (\%) $\approx$ Maj. Vote}} \\
 &  & \textbf{Samp.} & \textbf{DCR(S)} & \textbf{DCR(M)} & \textbf{Samp.} & \textbf{DCR(S)} & \textbf{DCR(M)} \\
\cmidrule(lr){3-5}\cmidrule(lr){6-8}
\midrule
 MATH500 & GPT-OSS & 83.9 & 84.6 & \textbf{85.2} & 85.0 & \textbf{85.8} & 85.4 \\
 MATH500 & Llama-4 & 36.4 & 73.7 & \textbf{79.1} & 35.8 & 74.0 & \textbf{79.0} \\
 MATH500 & Llama-3.3 & 20.8 & 51.4 & \textbf{72.9} & 18.8 & 52.4 & \textbf{72.8} \\
 MATH500 & Granite-4 & 26.2 & 53.3 & \textbf{73.6} & 26.0 & 54.6 & \textbf{74.6} \\
\midrule
 AIME 2024 & GPT-OSS & 74.3 & \textbf{88.1} & 84.0 & 76.7 & \textbf{90.0} & 83.3 \\
 AIME 2024 & Llama-4 & 5.1 & 29.3 & \textbf{67.1} & 3.3 & 30.0 & \textbf{66.7} \\
 AIME 2024 & Llama-3.3 & 13.2 & 23.2 & \textbf{62.4} & 16.7 & 23.3 & \textbf{63.3} \\
 AIME 2024 & Granite-4 & 0.1 & 7.5 & \textbf{67.5} & 0.0 & 6.7 & \textbf{70.0} \\
\midrule
 AIME 2025 & GPT-OSS & 74.1 & \textbf{87.2} & 80.1 & 80.0 & \textbf{90.0} & 83.3 \\
 AIME 2025 & Llama-4 & 0.7 & 14.9 & \textbf{57.9} & 0.0 & 13.3 & \textbf{56.7} \\
 AIME 2025 & Llama-3.3 & 0.9 & 3.3 & \textbf{47.6} & 0.0 & 3.3 & \textbf{46.7} \\
 AIME 2025 & Granite-4 & 0.0 & 8.9 & \textbf{39.5} & 0.0 & 10.0 & \textbf{36.7} \\
\midrule
 MMLU-PRO & GPT-OSS & 79.3 & \textbf{81.4} & 80.2 & 80.0 & \textbf{81.4} & 81.3 \\
 MMLU-PRO & Llama-4 & 64.7 & 77.5 & \textbf{80.0} & 64.7 & 77.5 & \textbf{81.0} \\
 MMLU-PRO & Llama-3.3 & 68.9 & 73.5 & \textbf{77.6} & 69.7 & 73.5 & \textbf{77.7} \\
 MMLU-PRO & Granite-4 & 45.2 & 47.4 & \textbf{71.9} & 44.6 & 47.2 & \textbf{72.4} \\
\bottomrule
\end{tabular}
\end{table*}

\subsection{Evaluation Metrics}

Each method produces \(T\) stochastic trials per problem (typically \(T=25\)). A trial is marked correct if its final answer matches the ground truth. We report two metrics:

\begin{enumerate}
\item \textbf{Trial accuracy (attempt-level accuracy):} the fraction of correct trials over all problems and trials. If \(I\) is the number of problems, this is
\[
\mathrm{Acc}_{\text{trial}}
\;=\;
\frac{1}{IT}\sum_{i=1}^{I}\sum_{t=1}^{T}\mathbb{I}\!\left[a_{i,t}=a_i^\star\right].
\]
This captures average per-call success.

\item \textbf{Consistency (majority-correct rate):} the fraction of problems for which a strict majority ($>50\%$) of the method's trials are correct:
\[
\mathrm{Acc}_{\text{cons}}
\;=\;
\frac{1}{I}\sum_{i=1}^{I}\mathbb{I}\!\left[\frac{1}{T}\sum_{t=1}^{T}\mathbb{I}\!\left[a_{i,t}=a_i^\star\right] > 0.5\right].
\]

\textbf{Why this measures consistency:} Standard average accuracy can blur the distinction between ``lucky guesses'' and robust solutions. This metric captures \emph{reliability}: does the system's signal outweigh its noise? If a model is consistent on a problem (correctness $>0.5$), it effectively ``knows'' the solution and will converge to the right answer under majority voting. By contrast, a model that is only occasionally correct is unstable. Our main claim is that DCR improves consistency by transforming fragile, low-probability correct minority reports into stable, majority outcomes.
\end{enumerate}

\textbf{Connection to Baselines (Majority Vote and Best-of-$N$):} We use these metrics to bridge our results to standard community baselines without running separate experiments. \emph{Consistency} corresponds directly to the accuracy of a \textbf{Majority Vote} over the $T$ trials (since a consistent method yields a correct majority). \emph{Trial Accuracy} ($p$) proxies for \textbf{Best-of-$N$} (Pass@$N$) performance via the i.i.d.\ approximation $\text{Pass@}N \approx 1-(1-p)^N$. This allows us to demonstrate that DCR improves upon these baselines by showing gains in both reliability (Consistency) and coverage (Trial Accuracy).

\section{Results and Analysis}

Our results are organized around three core stories that align with our three contribution claims.

\subsection{Result 1: Single-round DCR amplifies correct minority reports}
We first analyze the single-round primitive. Table~\ref{tab:result1} presents the performance of single-round DCR compared to baseline Sampling.
Across all datasets and models, DCR consistently improves both trial accuracy and consistency (see Appendix~\ref{app:mmlu-category-table} for detailed breakdown by category on MMLU-PRO).
For example, on AIME 2024, GPT-OSS improves from 74.3\% trial accuracy (Sampling) to 88.1\% (DCR Single), and from 76.7\% consistency to 90.0\%.
Notable gains are also observed for weaker models; Llama-4 on MATH500 improves from 36.4\% to 73.7\% trial accuracy.
Critically, the \emph{Consistency} metric reveals that DCR transforms fragile, low-probability correct answers (where the model is correct in a minority of trials) into stable, majority outcomes.
Comparing DCR (Single) vs.\ DCR (Mix) shows that while mixing proposals can help weaker models (e.g., Llama-3.3 on AIME 2024 improves from 23.2\% to 62.4\% when reviewing a mixed pool), the strongest single-model performance often comes from self-correction (DCR Single) or strong-reviewer setups.
Crucially, we observe a distinct failure mode in GPT's DCR mix cases: weaker models' incorrect proposals can sometimes ``pollute'' the judgment of stronger models, causing them to hallucinate or drift from a correct path. This implies that mixing is most effective when lifting weak models toward a strong consensus, rather than diluting strong experts with weaker noise.

This ability to amplify correct minority reports suggests that reconciliation acts as a \emph{selection operator}, improving the signal-to-noise ratio of the solution pool. This observation motivates our next experiment: if one round improves the pool, can \emph{recursively} applying this operator drive the system toward stable agreement on the correct answer? We investigate this in the next section.

\subsection{Result 2: Recursive DCR enables attentive test-time compute allocation}
\label{sec:recursive_recovery}

\begin{figure*}[ht]
\begin{center}
\centerline{\includegraphics[width=0.95\textwidth]{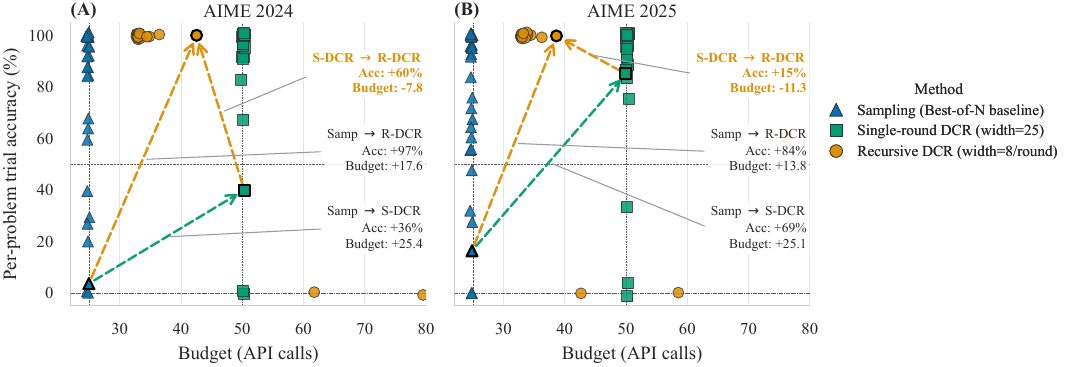}}
\caption{
\textbf{Accuracy versus test-time compute on (A) AIME 2024 and (B) AIME 2025.} Each dot corresponds to a single problem (points are jittered slightly to avoid overlap). The y-axis reports \textbf{per-problem trial accuracy} (fraction correct over repeated stochastic trials for that problem, in \%). The x-axis reports total API calls (exploration + reconciliation) as a proxy for budget spending. Colors indicate the aggregation method: Sampling (blue triangles, 25 calls), single-round DCR (green squares, 50 calls; reviewer width \(K=25\)), and recursive DCR (orange circles, variable calls; reviewer width \(K=8\) per round). Dashed arrows highlight the progression on a representative problem. Notably, the \textbf{S-DCR $\to$ R-DCR transition} (bold orange annotation) shows that recursive DCR achieves \emph{both} higher accuracy \emph{and} lower budget than single-round DCR---a ``win-win'' enabled by early stopping on easy problems, while preserving the ability to allocate more compute to persistent disagreements.
}
\label{fig:acc_vs_compute}
\end{center}
\vskip -0.25in
\end{figure*}

We next analyze the \emph{recursive} variant of DCR. Figure~\ref{fig:acc_vs_compute} plots per-problem trial accuracy (y-axis) against the total compute budget (x-axis, API calls).
Table~\ref{tab:result2_recursive} summarizes the efficiency gains.

\begin{table}[t]
\centering
\small
\setlength{\tabcolsep}{6pt}
\renewcommand{\arraystretch}{1.0}
\caption{\textbf{Result 2: Recursive DCR is more accurate and efficient.} compared to single-round DCR (budget=50). Recursive DCR stops early on easy problems (unanimity), reducing average compute by $\sim$27\%, while achieving higher accuracy.}
\label{tab:result2_recursive}
\begin{tabular}{llrr}
\toprule
\textbf{Dataset} & \textbf{Method} & \textbf{Acc. (\%)} & \textbf{Avg. Calls} \\
\midrule
\textbf{AIME 2024} & Single-round DCR & 88.1 & 50.0 \\
 & \textbf{Recursive DCR} & \textbf{93.3} & \textbf{36.4} \\
\midrule
\textbf{AIME 2025} & Single-round DCR & 87.2 & 50.0 \\
 & \textbf{Recursive DCR} & \textbf{92.0} & \textbf{36.1} \\
\bottomrule
\end{tabular}
\end{table}

\noindent\textbf{Higher Accuracy, Lower Cost.} Recursive DCR achieves the highest performance on the most difficult benchmarks, reaching \textbf{93.3\% trial accuracy on AIME 2024} and \textbf{92.0\% on AIME 2025}. Crucially, it achieves these gains while using \textbf{$\sim$27\% less compute on average} than the single-round baseline (approx.\ 36 calls vs.\ 50 calls; Table~\ref{tab:result2_recursive}).

\noindent\textbf{Attentive Allocation.} This ``better and cheaper'' result is driven by the unanimous-consent stopping rule. On easy problems (dense cluster of orange circles at low budget in Figure~\ref{fig:acc_vs_compute}), the system converges in a single round (total $\sim$33 calls), saving resources. On hard problems (dashed arrows), it automatically escalates, spending up to 80 calls to resolve persistent disagreements. This confirms that test-time compute should be allocated \emph{attentively} based on problem difficulty, rather than uniformly.

\noindent\textbf{Depth vs.\ Width.} These results suggest that for reasoning tasks, allocating compute to \emph{depth} (sequential rounds) is more effective than \emph{width} (more reviewers). While increasing width merely reduces variance via averaging, increasing depth allows the system to \emph{refine} its reasoning: later rounds condition on the structured disagreements surfaced by earlier rounds, effectively breaking deadlocks that a single-round consensus cannot resolve.
For instance, Appendix~\ref{app:recovery-example} details a case where the initial pool contained \emph{only} incorrect answers. In Round 2, two independent reviewers pinpointed why the leading ``1024'' proposal overcounts: maximality rules out any coloring that would force an empty row or column (equivalently, the row/column color sets must match), yielding the corrected count 902. This reasoning then persuaded the remaining reviewers in Round 3, leading to rapid unanimity. This ``reflective debugging'', using disagreement as a prompt to find the truth, is unique to the recursive setting.

\noindent\textbf{Comparison to Peer-Discussion.} The power of this ``critical review'' mechanism is evident when compared to peer-discussion baselines. We replicated the ReConcile round-table architecture \cite{chen2024reconcile} using the same models\footnote{We implemented the discussion structure and prompts from \citet{chen2024reconcile} but utilized our model ensemble to ensure a controlled comparison on the newer AIME benchmarks.}, achieving only 40.0\% (AIME 2024) and 53.3\% (AIME 2025)—nearly 40 points lower than DCR. This demonstrates that simply aggregating or debating peer opinions is insufficient; the explicit \emph{reviewer-author} distinction in DCR is necessary to overturn incorrect majority consensus.

\noindent\textbf{Stability.} We further observe that recursive DCR yields high result stability. Across 10 repeated runs per problem, the system typically either solves the problem in every trial (100\% consistency) or fails in every trial (0\%), with very few tasks showing high variance (e.g., only one problem in AIME 2025 had intermediate success rates). This indicates that the recursive mechanism is robust for problems within the model's capability frontier.

\subsection{Result 3: Dispersion as a Proxy for Relative Difficulty}
\label{sec:dispersion_diagnostic}

\begin{figure*}[ht]
\centering
\includegraphics[width=0.8\textwidth]{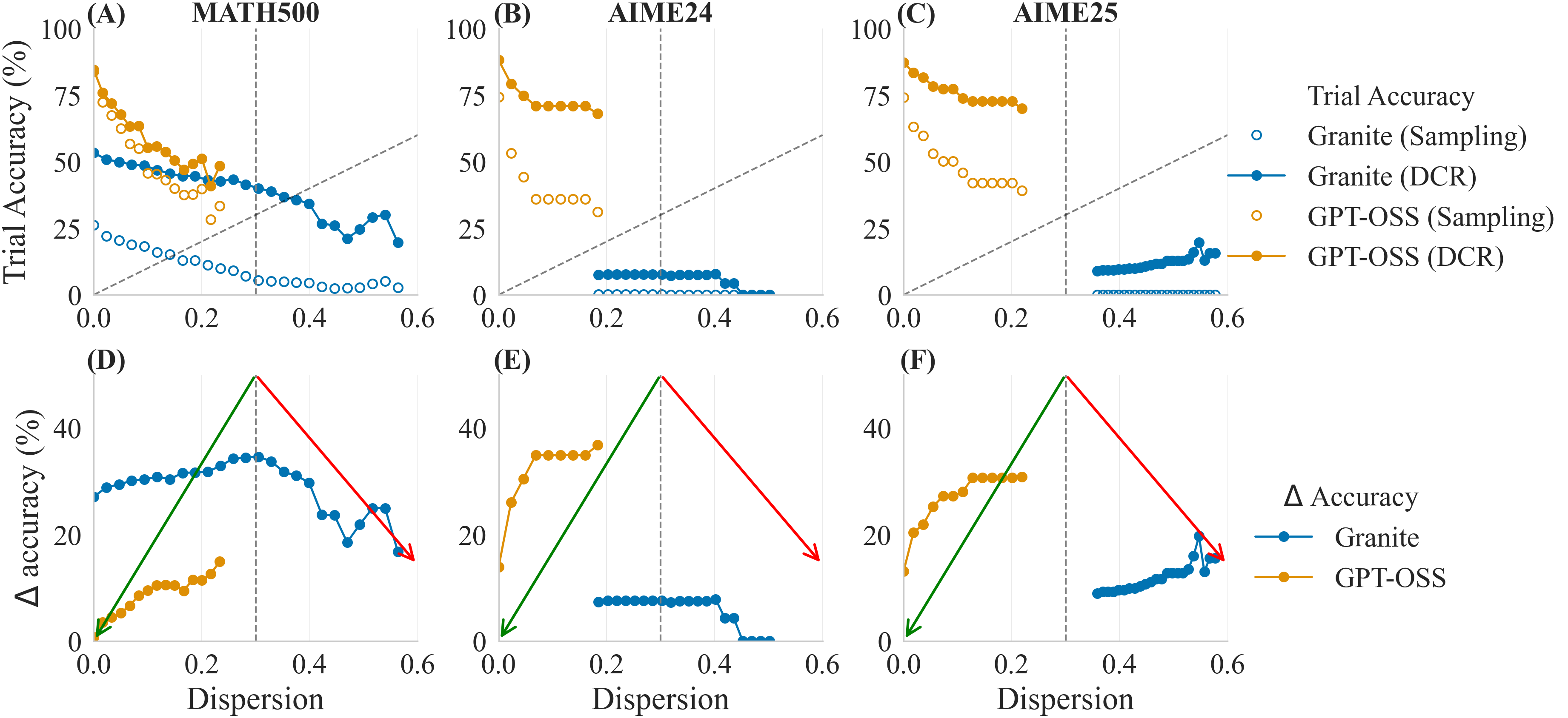}
\caption{
\textbf{Dispersion identifies the ``Sweet Spot'' for test-time compute.}
\textbf{Top row (A--C):} Trial accuracy for Sampling vs.\ DCR on the subset of tasks with dispersion $\ge d$. Low dispersion predicts high baseline accuracy; high dispersion signals difficulty.
\textbf{Bottom row (D--F):} The accuracy gain ($\Delta_{\text{DCR-Samp}}$) on the same subsets.
The \textcolor{green!60!black}{\textbf{green arrow}} marks the ``Recoverable Uncertainty'' regime, where filtering for higher dispersion isolates tasks where reconciliation is most effective.
The \textcolor{red!80!black}{\textbf{red arrow}} marks the ``Capability Collapse'' regime ($d \gtrsim 0.3$), where extreme dispersion indicates the model is overwhelmed, and gains diminish.
}
\label{fig:dispersion_vs_diagnostic}
\vskip -0.15in
\end{figure*}

Recursive DCR adaptively allocates compute based on disagreement. To understand \emph{when} this allocation is effective, we analyze \textbf{dispersion} (the geometric spread of a model's initial exploration samples) as a proxy for the \textbf{relative difficulty} of a task given a model's capabilities. Figure~\ref{fig:dispersion_vs_diagnostic} analyzes performance across dispersion thresholds.

\noindent\textbf{Dispersion Correlates with Difficulty (Top Row).}
The top panels show that dispersion is a strong, calibration-free signal of task difficulty. For both models, baseline accuracy (open circles) drops monotonically as dispersion increases. The dashed diagonal ($y=100d$) serves as an approximate separatrix: tasks with dispersion to the left of this line are typically within the model's capability frontier, while those to the right are effectively unsolvable noise.

\noindent\textbf{The ``Sweet Spot'' of Recoverable Uncertainty (Bottom Row).}
Crucially, the marginal benefit of DCR is not uniform. The bottom panels reveal three distinct regimes of test-time compute utility:

\begin{enumerate}
    \item \textbf{High Confidence ($d \approx 0$):} When dispersion is near-zero, the model is consistent. Reconciliation yields minimal gains because there is no disagreement to resolve.
    \item \textbf{Recoverable Uncertainty ($0 < d \lesssim 0.3$):} As tasks become harder, the model produces diverse but plausible reasoning paths. Here, DCR is most effective (positive slope, green arrow), acting as a selection pressure that identifies correct reasoning among the variants. We term this the \emph{Zone of Proximal Development} for the model.
    \item \textbf{Capability Collapse ($d \gtrsim 0.3$):} When dispersion becomes extreme, the model enters a regime of stochastic guessing. The accuracy gain from reconciliation flattens or degrades (red arrow), as the consensus mechanism struggles to extract a signal from high-entropy noise.
\end{enumerate}

\noindent\textbf{Implication for Deployment.}
This finding solves a key problem in test-time compute: knowing \emph{when} to scale. Dispersion serves as a cheap, unsupervised diagnostic. A system can aggressively reconcile tasks in the ``recoverable'' zone while flagging high-dispersion tasks for human review or escalation to a stronger model, rather than wasting compute on hallucinations.

\section{Discussion and Limitations}

\noindent\textbf{Positioning Relative to Verifier-Based Methods}
Our results demonstrate that DCR is effective in settings where ground-truth labels or trained verifiers are scarce. While verifier-based approaches are powerful when reward signals are available, DCR provides a complementary framework for \emph{verifier-free} environments, leveraging the intrinsic consistency of the solution space as a proxy for correctness.

\noindent\textbf{Efficiency via KV Cache Reuse}
A frequent critique of ensemble-based reasoning is the computational cost of repeated sampling. However, this cost is rapidly diminishing due to advances in inference system architecture. Techniques for efficient Key-Value (KV) cache reuse, such as LMCache \cite{Cheng2025LMCache}, CacheGen \cite{Liu2024CacheGen}, and CacheBlend \cite{Yao2025CacheBlend}, allow the heavy lifting of processing the prompt to be amortized across multiple samples. Furthermore, major cloud providers (e.g., OpenAI, Google Gemini) now natively support context caching, enabling repeated sampling of the same query with significantly reduced latency and cost. Consequently, the overhead of exploring divergent solution paths is becoming a negligible factor in modern deployment stacks.

\noindent\textbf{Mechanistic Intuition: Selection is Easier than Generation}
Across datasets, we observe that DCR often succeeds where majority voting fails (Result 1). This supports the intuition that \emph{selection} (recognizing a correct argument among candidates) is a simpler cognitive task than \emph{generation} (producing the argument from scratch). DCR transforms inference into an evolutionary process: the ``Divergent'' phase generates a population of diverse genotypes (reasoning traces), and the ``Convergent'' phase acts as a selection operator, filtering for robustness. Recursive application amplifies this selection pressure, allowing the correct answer to dominate even if it started as a minority.

\noindent\textbf{Dispersion as a Measure of Relative Difficulty}
A key deployment challenge is the ``Router Paradox'': estimating compute needs \emph{before} solving a query. While ``Predict-then-Route'' systems (e.g., vLLM Semantic Router~\cite{vllm_semantic_router2025}) work well on known distributions, they struggle to generalize to novel tasks. We advocate for ``Sample-then-Assess'': generating a small initial batch to measure dispersion provides a direct, training-free signal of relative difficulty. Specifically, low dispersion indicates high confidence (stop, or escalate if checking for bias); moderate dispersion signals a high-return zone for recursive compute; and extreme dispersion suggests the system is overwhelmed (abstain or flag for review). This provides a dynamic, self-calibrating compute budget that adapts to the model's capabilities without retraining a separate router.

\section{Future Work}

\textbf{Upfront Cost of Estimation.}
A limitation of using dispersion as a gating signal is the ``cold start'' cost: one must generate $N$ proposals (e.g., $N=25$ in our experiments) just to estimate it. In latency-sensitive production environments, this initial investment may be prohibitive. Future work could investigate adaptive probing strategies (e.g., starting with $N=5$ to estimate variance) or hybrid systems where a human expert or a lightweight heuristic decides which high-risk tasks warrant the full ``interview'' process of divergent exploration.

\textbf{Bias in Low-Dispersion Regimes.}
While low dispersion signals high confidence, it does not guarantee correctness. A model can be consistently wrong (``confident hallucination''). DCR is less effective here. Addressing this requires \emph{heterogeneous} ensembles (mixing proposals from different model families, as hinted in Experiment 2) to break the bias, rather than simply scaling compute on a single model.

\textbf{Stop Criteria Optimization.}
Our recursive system uses a strict unanimous-consent stopping rule. While robust, this may be overly conservative. Future research should explore ``soft consensus'' thresholds or probabilistic stopping rules to further optimize the trade-off between compute cost and accuracy.

\textbf{Multi-Signal Dispersion.}
Our current dispersion metric uses answer embeddings. Additional signals could include:
(i) \textbf{reasoning path dispersion}: disagreement in intermediate steps, not just 
final answers;
(ii) \textbf{confidence dispersion}: variance in model-reported confidence scores;
(iii) \textbf{token-level entropy}: uncertainty at the generation level.

\section{Conclusion}

We introduced Divergent--Convergent Reasoning (DCR), an ensemble-based inference primitive that transforms test-time compute from a uniform resource into an adaptive control variable. Across MATH500, AIME, and MMLU-PRO, we demonstrated that (i) single-round DCR amplifies correct minority reports where majority voting fails, (ii) recursive reconciliation achieves state-of-the-art accuracy with reduced compute by stopping early on consensus, and (iii) dispersion serves as a training-free diagnostic for task difficulty.

Beyond these empirical gains, our results suggest a fundamental shift in how we view model variance. Standard aggregation methods often treat disagreement as noise to be averaged away. In contrast, DCR reveals that disagreement is a high-value signal of \emph{recoverable uncertainty}. By structurally separating the generation of diverse hypotheses from their critical review, DCR emulates a ``System 2'' process—moving from rapid pattern matching to deliberate, self-correcting deliberation. This implies that the future of robust reasoning lies not merely in scaling parameters, but in scaling the \emph{structure} of inference: allowing systems to diverge in exploration so that they may reliably converge on the truth.

\section*{Impact Statement}

This work studies methods for allocating test-time computation in large language models without access to verifiers. Potential benefits include improved inference efficiency and better uncertainty awareness in deployed systems. Risks include high-confidence agreement on incorrect solutions; we explicitly analyze such failure modes and show that persistent disagreement can be used to trigger abstention or additional computation.

\bibliography{main}
\bibliographystyle{icml2026}

\newpage
\appendix
\onecolumn

\section{Differentiation from Prior Work (Supplementary)}
\label{app:prior-art-diff}

\subsection{Verifier-free (VF) post-training and trace imitation}

\paragraph{\textbf{s1: Simple test-time scaling} \cite{muennighoff2025s1}.}
\textbf{What it does.} s1 targets test-time scaling via a lightweight recipe: supervised fine-tuning on a small curated set of reasoning traces and a simple test-time technique (“budget forcing”) that increases or decreases the length of the model’s thought process.
\textbf{How DCR differs.} DCR is \emph{training-free at deployment}: it does not require collecting or distilling expert traces to teach a single model to “think longer.” Instead, DCR uses \emph{inter-agent dispersion} across independently generated proposals as a proxy for uncertainty, and allocates compute by invoking a \emph{convergent reconciliation} step that resolves disagreements across solutions.
\textbf{Key differences.}
\begin{itemize}
  \item \textbf{Compute control signal:} s1 controls compute directly (forced lengthening/termination) on a \emph{single trajectory}; DCR uses \emph{cross-solution disagreement} to decide when to spend additional compute on \emph{reconciliation}.
  \item \textbf{Where compute goes:} s1 spends more tokens on extended single-model deliberation; DCR spends compute on generating diverse proposals and a reviewer-style reduce stage.
  \item \textbf{Artifacts required:} s1 depends on curated supervision for post-training; DCR operates without new training data or reward models.
\end{itemize}
\textbf{Scope caveat.} DCR does not claim to replace post-training approaches; rather it targets the common deployment regime where a reliable verifier is unavailable and re-training is impractical.

\paragraph{\textbf{SFT memorizes, RL generalizes} \cite{chu2025sft}.}
\textbf{What it does.} This line of work studies the comparative behavior of supervised fine-tuning (SFT) and reinforcement learning (RL) in terms of memorization and generalization, largely as a \emph{post-training} analysis of learning dynamics.
\textbf{How DCR differs.} DCR is an \emph{inference-time} multi-agent procedure; it allocates test-time compute based on disagreement across generated solutions, without relying on post-training to shape the model’s internal reasoning behavior.
\textbf{Key differences.}
\begin{itemize}
  \item \textbf{Goal:} post-training generalization properties vs inference-time compute allocation and reconciliation.
  \item \textbf{Mechanism:} parameter updates vs structured generation + aggregation.
\end{itemize}
\textbf{Scope caveat.} Our VF framing concerns test-time decision-making without a verifier, not a claim about SFT/RL tradeoffs in training.

\paragraph{\textbf{LIMO: Less is More for Reasoning} \cite{ye2025limo}.}
\textbf{What it does.} LIMO argues that strong reasoning can be elicited with a very small number of carefully curated demonstrations, using simple supervised fine-tuning to provide “cognitive templates.”
\textbf{How DCR differs.} DCR does not depend on eliciting reasoning via demonstrations; instead it improves performance by \emph{structuring inference} into divergent proposal generation followed by reviewer-style reconciliation, using disagreement as the control signal for allocating more compute.
\textbf{Key differences.}
\begin{itemize}
  \item \textbf{Training dependence:} LIMO improves a base model via SFT; DCR improves outputs via multi-agent inference-time reconciliation.
  \item \textbf{Uncertainty signal:} LIMO does not require cross-solution dispersion as a controller; DCR explicitly uses dispersion/consistency to decide whether to continue.
\end{itemize}

\subsection{Early stopping / test-time compute allocation}

\paragraph{\textbf{Stop When Enough} \cite{sun2025stop}.}
\textbf{What it does.} Uses reflection cues (self-check, strategy shift, uncertainty expression, retrospective revisions) combined with semantic redundancy (cosine similarity to previous steps in the chain) to detect when to halt generation within a single reasoning trajectory.
\textbf{How DCR differs.} DCR uses \emph{inter-agent dispersion} across independently generated proposals as the control signal, allocating \emph{more} compute for reconciliation when disagreement is high, rather than stopping early within a single trajectory. DCR operates at the multi-agent level rather than the intra-trajectory level.
\textbf{Key differences.}
\begin{itemize}
  \item \textbf{Signal:} Intra-trajectory reflection + semantic redundancy vs inter-agent answer dispersion.
  \item \textbf{Decision policy:} Early stopping within single chain vs allocating more compute for multi-agent reconciliation.
  \item \textbf{Granularity:} Single trajectory monitoring vs cross-proposal analysis.
\end{itemize}

\paragraph{\textbf{HALT-CoT} \cite{laaouach2025haltcot}.}
\textbf{What it does.} Monitors entropy of the model's probability distribution over candidate answers after each CoT step, halting generation when entropy drops below a threshold (indicating high confidence in the answer).
\textbf{How DCR differs.} DCR uses \emph{inter-agent disagreement} as the signal rather than intra-trajectory answer entropy, and responds to high dispersion by allocating \emph{more} compute for reconciliation rather than stopping generation.
\textbf{Key differences.}
\begin{itemize}
  \item \textbf{Signal:} Answer entropy within single trajectory vs answer dispersion across multiple agents.
  \item \textbf{Decision policy:} Stop when confident (low entropy) vs allocate more compute when uncertain (high dispersion).
  \item \textbf{Uncertainty source:} Model's internal confidence vs disagreement between independent agents.
\end{itemize}

\subsection{Multi-agent debate, judging, and aggregation}

\paragraph{\textbf{Improving factuality and reasoning in language models through multiagent debate} \cite{du2023improving}.}
\textbf{What it does.} Proposes a multi-agent debate procedure in which multiple replicas of an LLM generate initial responses, then iteratively update their responses over multiple rounds after seeing other agents' responses (typically via concatenation/summarization prompts), empirically improving both reasoning and factuality.
\textbf{How DCR differs.} DCR separates \emph{divergent exploration} (independent proposal generation) from an explicit \emph{convergent reconciliation} stage that is instructed to identify consensus steps and pinpoint divergent steps. Unlike standard debate, which often uses a fixed number of rounds, Recursive DCR uses \emph{unanimous-consent stopping} to dynamically allocate compute: it continues reconciling only while disagreement persists.
\textbf{Key differences.}
\begin{itemize}
  \item \textbf{Structure:} Iterative response revision rounds vs explicit divergent-then-reconciliation pipeline with reviewer-style reduce.
  \item \textbf{Control signal:} Debate duration often fixed/heuristic vs unanimous-consent stopping.
  \item \textbf{Reconciliation primitive:} Agent self-revision conditioned on others vs explicit reconciliation over disagreements into a new coherent solution.
\end{itemize}

\paragraph{\textbf{ReConcile} \cite{chen2024reconcile}.}
\textbf{What it does.} Multi-round discussion framework where diverse LLMs engage in iterative debate to convince each other, using confidence estimation and convincing samples, with final aggregation via weighted voting based on recalibrated confidence scores.
\textbf{How DCR differs.} DCR adopts a \textbf{Reviewer-Author} framework rather than a \textbf{Peer-Discussion} framework. Instead of agents trying to convince each other (peer persuasion), DCR agents act as neutral reviewers synthesizing a new answer from a set of proposals. Furthermore, DCR uses a strict \emph{unanimous-consent} stopping rule to ensure reliability, whereas ReConcile relies on confidence-weighted voting.
\textbf{Key differences.}
\begin{itemize}
  \item \textbf{Mechanism:} Peer persuasion (debate) vs Reviewer synthesis (reconciliation).
  \item \textbf{Stopping Rule:} Debate until consensus/max-rounds vs Unanimous-Consent Stopping.
  \item \textbf{Outcome:} Weighted vote of existing agents vs Generation of a new reconciled solution.
\end{itemize}

\paragraph{\textbf{Multi-Agent Debate for LLM Judges with Adaptive Stability Detection} \cite{hu2025multiagent}.}
\textbf{What it does.} Uses a time-varying Beta-Binomial mixture model with Kolmogorov-Smirnov testing to detect when judge consensus distributions stabilize, halting the debate process when convergence is achieved.
\textbf{How DCR differs.} DCR measures dispersion across independently generated solution proposals rather than judge consensus distributions, and responds to high dispersion by allocating compute to reconciliation rather than stopping the process.
\textbf{Key differences.}
\begin{itemize}
  \item \textbf{Signal:} Judge consensus distribution stability vs solution proposal dispersion.
  \item \textbf{Context:} Judging/evaluation task vs problem-solving task.
  \item \textbf{Response to signal:} Stop debate vs allocate more compute for reconciliation.
\end{itemize}

\paragraph{\textbf{LLM-TOPLA} \cite{tekin2024llmtopla}.}
\textbf{What it does.} Diversity-optimized ensemble method that uses a focal diversity metric to prune sub-ensembles from a pool of LLMs and employs learn-to-ensemble approaches to resolve output inconsistencies among component models.
\textbf{How DCR differs.} DCR uses disagreement as a \emph{controller} for allocating additional compute to generate new synthesized solutions, rather than only selecting and weighting existing model outputs. DCR creates novel solutions through metacognitive analysis.
\textbf{Key differences.}
\begin{itemize}
  \item \textbf{Goal:} Optimal selection/weighting of existing outputs vs reconciliation of new solutions.
  \item \textbf{Compute allocation:} Static ensemble construction vs dynamic compute allocation based on dispersion.
  \item \textbf{Output generation:} Weighted combination of existing predictions vs construction of new coherent solutions.
\end{itemize}

\paragraph{\textbf{Beyond Majority Voting: LLM Aggregation by Leveraging Higher-Order Information} \cite{ai2025beyond}.}
\textbf{What it does.} Improves on majority voting in an \emph{unsupervised} aggregation setting by leveraging higher-order (in particular, second-order) information about \emph{dependencies between agents}---e.g., empirical co-occurrence/conditional distributions of predictions across agents (after random shuffling of option labels), which can capture correlation structure beyond marginal accuracies.
\textbf{How DCR differs.} DCR is not a label-free weighting rule over one-shot predictions; it is a two-phase inference procedure that uses disagreement as a signal to allocate compute for reconciliation, and generates new solutions through structured analysis rather than only reweighting existing answers.
\textbf{Key differences.}
\begin{itemize}
  \item \textbf{Information use:} Higher-order dependence information for aggregation vs dispersion for compute allocation.
  \item \textbf{Output:} Weighted combination of existing answers vs reconciliation of new coherent solutions.
  \item \textbf{Compute control:} Static aggregation vs dynamic allocation based on disagreement level.
\end{itemize}

\subsection{Minority/outlier theory support}

\paragraph{\textbf{Promoting Erroneous Divergent Opinions Increases the Wisdom of Crowds} \cite{barrera2024divergent}.}
\textbf{What it does.} This work provides theoretical/empirical support (in human-crowd settings) for the idea that divergent opinions can improve aggregate outcomes under certain conditions.
\textbf{How DCR relates.} We cite this as \emph{motivation} for treating disagreement as a valuable signal rather than noise. DCR operationalizes this intuition for LLM systems by detecting dispersion and invoking a reconciliation stage that explicitly analyzes points of disagreement.
\textbf{Scope caveat.} This is not an LLM method; we do not claim it directly evaluates our system, only that it supports the plausibility of leveraging divergence.

\section{Prompts}
\label{app:prompts}

\subsection{Divergent Exploration Prompt (Sampling)}

\begin{WrappedVerbatim}
You are an expert mathematician tasked with solving a complex math problem. Your goal is to find the correct solution by working through the problem systematically.

Follow these steps to solve the problem:

1. UNDERSTAND THE PROBLEM
   - Restate the problem in your own words (briefly) to ensure comprehension
   - Identify what you are asked to find
   - Note any constraints, edge cases, or hidden conditions
   - Define all variables and symbols you will use

2. PLAN YOUR APPROACH
   - Identify the mathematical concepts and techniques needed
   - Consider multiple solution strategies if applicable
   - State any key lemmas you intend to prove along the way
   - Choose the most efficient approach

3. EXECUTE THE SOLUTION
   - Work through the problem step-by-step
   - Show all your work clearly
   - Provide proofs for intermediate claims when non-trivial
   - Use proper mathematical notation
   - Perform calculations carefully and symbolically where possible
   - Double-check arithmetic or algebraic manipulation at each stage

4. VERIFY YOUR ANSWER
   - Substitute your result back into the original conditions to confirm validity
   - Check boundary or special cases that could invalidate the solution
   - If multiple solutions are possible, justify why only your answer satisfies all requirements
   - Optionally, solve via an alternative method as a cross-check

5. PRESENT YOUR FINAL ANSWER
   - State your final answer clearly
   - Include units if applicable
   - If numeric, give exact value; if expression, present fully simplified
   - Ensure the answer format matches what the problem requests

Output specification:
Respond with structured JSON output containing your step-by-step reasoning and final answer.
\end{WrappedVerbatim}

\subsection{Convergent Reconciliation Prompt}

\begin{WrappedVerbatim}
You are an expert mathematician reviewing several proposed solutions to the same math problem. Use these proposals as clues to guide you toward the single correct answer.

KEY INSIGHT: Areas where solutions disagree are likely the hardest parts of the problem and require your deepest analysis.

Instructions:
1. Mark any step that appears in all solutions as a ``consensus step''—it is probably correct and needs only brief verification.
2. Pinpoint the exact area(s) where the solutions disagree—these ``divergent steps'' are the hardest parts of the problem and need thorough analysis.

Analysis Process:
1. Treat the divergent steps as way-points in the solution space. Compare the alternative approaches and decide which one is mathematically sound.
2. Explain (to yourself in your <thinking>) why the incorrect paths fail and how the correct path resolves the conflict.

Retain the consensus steps that you have verified as correct.

Replace each divergent step with the correct reasoning you have identified (or your own fresh reasoning if none are correct).

Ensure the overall solution flows logically from premises to conclusion.

Output specification:
Respond with structured JSON output containing your step-by-step reasoning and final answer.
\end{WrappedVerbatim}

\section{Dispersion metric}
\label{app:dispersion}
In Result~3, we use \emph{dispersion} as a post-hoc diagnostic of the stability of a model’s exploration-stage outputs for a given problem. High dispersion indicates that repeated sampling yields qualitatively different reasoning paths and answers (i.e., the model explores a broad solution space), whereas low dispersion indicates that sampling collapses to essentially the same answer across runs, which may be either correct or systematically biased.

Formally, let $\{y_n\}_{n=1}^{N}$ denote the candidate answers produced during the exploration stage for a given problem. Each answer is embedded using a fixed sentence embedding model $\phi(\cdot)$ that maps variable-length text to a 1024-dimensional vector (Sentence-BERT; \cite{reimers2019sentence}, with a RoBERTa backbone \cite{liu2019roberta}). We compute the centroid
\[
\bar{z} = \frac{1}{N}\sum_{n=1}^{N} \phi(y_n),
\]
and define dispersion as the average squared Euclidean distance to this centroid:
\[
D_{\text{sampling}} = \frac{1}{N}\sum_{n=1}^{N} \left\lVert \phi(y_n) - \bar{z} \right\rVert_2^2 .
\]

This centroid-based measure is robust to degenerate majority effects, in which a tight but incorrect cluster dominates a small number of diverse (and potentially correct) outlier answers.

\section{Result example: Emergence of the correct solution from incorrect proposals}
\label{app:recovery-example}
\noindent\textbf{Summary of Recovery Mechanism.} This example illustrates how Recursive DCR recovers the correct answer (902) even when the initial exploration phase produces exclusively incorrect proposals (Round 1). In the traces below, observe how two reviewers in Round 2 (Reviewers 1 and 3) independently engage in \textbf{reflective debugging}: they identify the precise failure mode in the dominant ``1024'' argument. The mistake is \emph{not} arithmetic, but a missing feasibility constraint: maximality forbids configurations that would leave any row or column empty, which forces the \emph{set of colors appearing among rows} to match the \emph{set of colors appearing among columns}. This yields the corrected count \(1+1+(2^5-2)^2 = 902\). In Round 3, this logic persuades the remaining reviewers---with Reviewer 4 explicitly reframing the fix as excluding 122 invalid colorings, i.e.\ $1024 - 122 = 902$---leading to immediate unanimous convergence.

\noindent\textbf{Setup.} Dataset: AIME24, problem \texttt{aime24-2024-II-9}. Model: \texttt{gpt-oss-120b}. Reviewer width \(K=4\). Repeat: 1. Ground-truth answer: 902.
\textbf{Evolution of Consensus:} We show how the correct answer emerges from 0\% in Round 1 to 100\% unanimity in Round 3.
\begin{table}[ht]
\centering
\caption{Evolution of consensus over 3 rounds. Columns show the answer produced by each of the 4 reviewers in each round. The correct answer (\textbf{902}) is highlighted in bold. Starting from a completely incorrect state in Round 1, the correct answer emerges in Round 2 (discovered independently by Reviewers 1 and 3) and persuades the remaining reviewers to reach unanimity in Round 3.}
\label{tab:recovery_evolution}
\begin{tabular}{c|cccc|l}
\toprule
\textbf{Round} & \textbf{Reviewer 1} & \textbf{Reviewer 2} & \textbf{Reviewer 3} & \textbf{Reviewer 4} & \textbf{Consensus State} \\
\midrule
1 & 1024 & 13806 & 1024 & 122 & Diverged (All Incorrect) \\
2 & \textbf{902} & 1024 & \textbf{902}  & 1024 & \textbf{Emergence} (Correct Minority) \\
3 & \textbf{902} & \textbf{902} & \textbf{902} & \textbf{902} & \textbf{Unanimity} (Stopped) \\
\bottomrule
\end{tabular}
\end{table}

\subsection{Output and reasoning traces}
\noindent\textbf{Round 1 Reviewer 1 (answer = 1024, incorrect).}
\begin{small}
\begin{quote}
We model the placement by assigning a colour (white or black) to each of the 5 rows and each of the 5 columns. A chip can be placed only at the intersection of a row and a column that have the same colour; the chip's colour is then forced to be that colour. Maximality means that no empty cell can be filled without breaking the colour rule. If a row were empty while some column contained a chip, the cell at their intersection could be filled with a chip of the column's colour, contradicting maximality. Hence an empty row cannot coexist with a non-empty column, and similarly an empty column cannot coexist with a non-empty row. Therefore \textbf{every row and every column must contain at least one chip}; i.e. there are no empty rows and no empty columns.

With all rows and columns non-empty, each row can be chosen independently to be either white or black (2 choices), and the same for each column (2 choices). Once the colour sets are chosen, maximality forces that every intersection of a white row with a white column be occupied by a white chip, and every intersection of a black row with a black column be occupied by a black chip; intersections of rows and columns of opposite colours remain empty and cannot be filled without violating the colour condition.

The number of colour assignments is therefore:
\[
2^{5}\times 2^{5}=2^{10}=1024.
\]
All such assignments respect the chip supply limits (the most chips of one colour used is $5\times5=25$). Each assignment yields a distinct placement, so the total number of maximal placements is 1024.

All other proposed counts (e.g., 122, 902, 253, 254, etc.) arise from incorrect handling of the maximality condition---specifically, they allow configurations with empty rows and columns simultaneously, which would not be maximal because an additional chip could be placed at their intersection.
\end{quote}
\end{small}

\noindent\textbf{Round 1 Reviewer 2 (answer = 13806, incorrect).}
\begin{small}
\begin{quote}
We model the placement by assigning each row and each column a colour: white (W), black (B), or empty (E). A chip can be placed in a cell only when the row and column colours agree and are not empty; then the cell must be occupied, otherwise the configuration would not be maximal.

Maximality also forces that there is no pair consisting of an empty row and an empty column, because the intersection of such a pair would be an empty cell where a chip could be added without breaking the colour condition. Hence any admissible configuration must satisfy:
\[
\text{(no empty rows)} \;\; \text{OR} \;\; \text{(no empty columns)}.
\]

Furthermore, if a colour appears among the rows it must also appear among the columns (otherwise rows of that colour could not have any chip), and vice-versa. Thus the set of colours present among rows equals the set of colours present among columns.

We count assignments of colours to rows and columns satisfying the above.

\textbf{1. Count assignments with no empty rows.}
Each of the 5 rows can be W or B (2 choices), so $2^5 = 32$ row colourings.
\begin{itemize}
\item If all rows are W (1 case) then columns may be W or E (2 choices each), giving $2^5 = 32$ assignments.
\item If all rows are B (1 case) similarly 32 assignments.
\item If the rows contain both colours (30 cases) then each column may be W, B, or E (3 choices each), giving $3^5 = 243$ assignments per row pattern.
\end{itemize}
Thus
\[
N_{\text{rows}} = 1\cdot 32 + 1\cdot 32 + 30\cdot 243 = 7354.
\]

\textbf{2. Count assignments with no empty columns.} By symmetry the same number:
\[
N_{\text{cols}} = 7354.
\]

\textbf{3. Subtract the double-counted configurations where there are neither empty rows nor empty columns.}
Here every row and every column is coloured (no E). The colour sets must match, giving three possibilities:
\begin{itemize}
\item all rows and columns white (1)
\item all rows and columns black (1)
\item both colours appear. There are $(2^5-2)=30$ ways to colour the rows with both colours and likewise 30 ways for the columns, giving $30\cdot 30 = 900$.
\end{itemize}
Hence
\[
N_{\text{both}} = 1 + 1 + 900 = 902.
\]

\textbf{4. Inclusion-exclusion:}
\[
N_{\text{total}} = N_{\text{rows}} + N_{\text{cols}} - N_{\text{both}} = 7354 + 7354 - 902 = 13806.
\]

Each such assignment determines a unique maximal placement: a cell is occupied exactly when its row and column colours are equal (and not empty). The number of white chips is $|R_W|\cdot|C_W| \le 25$ and similarly for black, so the supply of 25 of each colour is never exceeded.

Therefore the number of ways to place the chips satisfying all conditions is \textbf{13806}.
\end{quote}
\end{small}

\noindent\textbf{Round 1 Reviewer 3 (answer = 1024, incorrect).}
\begin{small}
\begin{quote}
Let each row be assigned a colour (white, black, or empty) and each column similarly. If a row has a colour, any empty column intersecting it would allow adding a chip of that row's colour, contradicting maximality; thus no empty columns can exist when any row is coloured. The same argument forces that any empty row implies all columns are empty, which yields the completely empty grid, not maximal. Therefore in any maximal configuration there are no empty rows and no empty columns; every row and column is coloured either white or black.

With all rows and columns coloured, a chip can be placed in a cell $(i,j)$ iff the colour of row $i$ equals the colour of column $j$; otherwise the cell must stay empty, otherwise the colour condition would be violated. Maximality then forces that every such matching-colour cell is occupied (otherwise we could add a chip there). Hence a maximal placement is uniquely determined by the colour assignments to the 5 rows and 5 columns.

Each of the 5 rows can be white or black (2 choices) and each of the 5 columns can be white or black (2 choices), giving $2^5\cdot 2^5 = 2^{10} = 1024$ possible colour assignments. For any assignment the number of white chips is $(\#\text{white rows})(\#\text{white columns}) \le 5\cdot 5 = 25$ and similarly for black chips, so the supply of 25 of each colour is never exceeded. Every assignment therefore yields a valid maximal placement, and distinct assignments give distinct placements.

Thus the total number of ways to place the chips satisfying all conditions is 1024.
\end{quote}
\end{small}

\noindent\textbf{Round 1 Reviewer 4 (answer = 122, incorrect).}
\begin{small}
\begin{quote}
We model the placement by the sets of rows $R$ and columns $C$ that contain chips. Maximality forces that every intersection of a row in $R$ with a column in $C$ must be occupied, otherwise we could add a chip there without breaking the uniform-colour condition. Hence the occupied cells form the full rectangle $R\times C$, all of the same colour (white or black).

If there were an empty row $i$ ($i\notin R$) and an empty column $j$ ($j\notin C$), the cell $(i,j)$ would be empty but we could place a chip of either colour there, contradicting maximality. Therefore at least one of the two sets must be the whole set of rows or columns. So either $R$ is all 5 rows or $C$ is all 5 columns (or both).

\textbf{Case 1: $R$ = all 5 rows.} Choose any non-empty subset of columns $C$ (size $r = 1,\dots,5$). There are $\sum_{r=1}^{5}\binom{5}{r} = 2^{5}-1 = 31$ choices. For each choice the rectangle is monochromatic, and we may colour it white or black (2 choices). Hence $31\cdot 2 = 62$ configurations.

\textbf{Case 2: $C$ = all 5 columns.} Symmetrically we obtain another 62 configurations.

The configuration where both $R$ and $C$ are all rows and columns (the full $5\times5$ grid) has been counted in both cases. It can be all white or all black, i.e. 2 configurations, so we subtract one copy of these 2.

Total maximal placements $= 62 + 62 - 2 = 122$.

All configurations respect the chip supply (the rectangle uses at most 25 chips of its colour). The empty placement is not maximal because a chip could be added. Thus the required number of ways is 122.
\end{quote}
\end{small}

\noindent\textbf{Reasoning path (Round 2 Reviewer 1; answer = 902, correct; recovery).}
\begin{small}
\begin{WrappedVerbatim}
We need to determine correct answer. Let's analyze problem: 25 white indistinguishable chips, 25 black chips, 5x5 grid, at most one chip per cell. All chips in same row and same column have same colour. So for any row that contains any chips, all chips in that row must be same colour. Similarly for any column that contains any chips, all chips in that column must be same colour. Additionally maximal: cannot add any additional chip without violating either condition.

We need count of maximal placements.

We need to reason.

Let rows R_i, columns C_j. Let a row be ``colored'' if it contains at least one chip; then all chips in that row must be same colour, call row colour. Similarly column colored if contains at least one chip, column colour.

If a cell (i,j) is occupied, its colour must equal both row i colour and column j colour, so row colour = column colour. Therefore any occupied cell lies at intersection of a row and column of same colour.

Conversely, if a row i is coloured white and column j is coloured white, then cell (i,j) could be occupied white. Maximality says if both row and column are coloured and same colour, then cell must be occupied; otherwise we could add chip there without breaking conditions (since row and column already white, adding white chip maintains uniformity). So maximal placements correspond to assignments of colours (white/black) to some subset of rows and columns, such that:

- No row or column is both coloured white and black obviously; each coloured row has a colour.
- If a row is coloured (has chips) and a column is coloured of same colour, then cell occupied.
- If a row is coloured white and column coloured black (different colours) then cell must be empty, because cannot place chip of either colour (would violate one of row/column uniformity). That's okay; does it affect maximality? Could we add a chip of either colour? No because row expects white only, column expects black only, conflict, so can't add any chip. So maximality doesn't require filling those cells.

- If a row is coloured (has chips) and a column is uncoloured (empty), then cell (i,j) is empty. Could we add a chip? Row expects its colour, column currently has no chips so no colour restriction yet. Adding a chip of row's colour would give column a colour (that colour) and would be consistent (since column would then have all chips of that colour). So maximality would be violated unless column is also coloured with that colour already (i.e., column must be coloured). Therefore any coloured row forces all columns to be coloured (with same colour?) Wait: If row i is white, column j is currently empty (no chips). Adding a white chip at (i,j) would give column j a white chip, making column j coloured white. This would not violate any condition because column j would then have all chips white (maybe other rows also white). So maximality requires that there are no empty columns when there exists a coloured row. Similarly, any coloured column forces all rows coloured.

Thus either there are no coloured rows (i.e., no chips at all) or there are no empty columns. But empty rows also cause symmetry. Let's formalize:

Suppose there exists at least one chip, thus there is at least one coloured row and at least one coloured column. Then consider any row that is coloured (has chips). For any column that is currently empty (no chips), the intersection cell could be filled with chip of the row's colour, giving that column chips of that colour, which would be allowed. Therefore maximality forces that there are no empty columns. Similarly, any coloured column forces no empty rows. Hence when there is any chip, all rows and all columns must be coloured (i.e., contain at least one chip). So no empty rows or columns.

Thus maximal configurations correspond to assignments of colours (white/black) to each of the 5 rows and each of the 5 columns, with both sets fully assigned (no empties). Then the occupied cells are exactly those where row colour = column colour. Maximality ensures those are occupied; cells where colours differ remain empty.

Now need to ensure chip supply constraints: number of white chips = (#white rows)*(#white columns) <= 25 (since max 5*5). Similarly black chips <=25. Since each count is product of numbers <=5, max is 25, so always satisfied. Also total white chips cannot exceed 25, same for black; fine.

Thus number of configurations = 2^5 choices for row colours * 2^5 for column colours = 2^10 = 1024.

\end{WrappedVerbatim}

\noindent\textbf{Reflective Debugging (The ``Aha!'' Moment):}
\begin{WrappedVerbatim}
But need to consider possibility of all rows and columns coloured but maybe some colour not used? That's fine; e.g., all rows white, all columns white gives full grid white (25 chips) which uses all white chips but still within supply. All rows white, all columns black would give no matching colour because rows white columns black differ; then no cell occupied? Wait condition: cell occupied when row colour = column colour. If rows white, columns black, then no matches, so zero chips placed. But would that be maximal? If there are no occupied cells, but rows are coloured white (meaning they contain chips) contradictory because a row coloured white must have at least one chip. Our definition: a row is coloured if it contains at least one chip. In this assignment, rows white but columns black, there would be no intersections with same colour, so no chip placed, violating that rows contain chips. Therefore such assignments are not realizable because row colour assignment must be consistent with existence of chips in that row. So we must ensure that for each coloured row, there is at least one column of same colour, otherwise row would have zero chips. Similarly for each column.

Thus need to ensure that each row colour appears among columns as well (i.e., for each colour present in rows, there is at least one column of same colour). Similarly each column colour appears among rows. But if we assign colours arbitrarily, could we have a row colour that doesn't appear among columns? Example: all rows white, columns mixture of white and black, but at least one white column exists, then each white row will have chips at intersection with white columns (nonempty). So each row will have chips because there is at least one white column. So row colour appears among columns ensures each row gets at least one chip. Conversely, column colour appears among rows ensures each column gets at least one chip.

If a colour appears only in rows but not columns, then rows of that colour would have zero chips, contradicting that they are coloured (i.e., contain chip). However could we consider that row being actually empty (no chips) and thus not coloured? But earlier we argued no empty rows allowed when any chip present. So such assignment invalid.

Thus the valid assignments are those where the set of colours used in rows equals the set used in columns, and each colour set is nonempty (since there is at least one chip). Could we have only one colour overall? Yes, all rows white and at least one column white (others could be white too). But if any column is black while rows all white, then black column would have no matching row white, so that column would be empty, contradicting no empty columns. So all columns must also be white. So the all-white assignment is valid (full white grid). Similarly all-black.

Thus the only valid assignments are those where rows and columns are coloured with the same colour set, and each colour present appears in both rows and columns.

Let's count such assignments.

Let possible colour sets be: {W}, {B}, {W,B}. For set {W}: all rows white, all columns white. That's 1 assignment.

Set {B}: all rows black, all columns black. 1 assignment.

Set {W,B}: rows can be any nonempty subset of white rows and black rows, but both colours must appear among rows, i.e., at least one white row and at least one black row. Similarly columns must have at least one of each colour. Number of ways: (2^5 - 2) = 30 ways to assign colours to rows with both colours present. Same for columns: 30 ways. Total 30*30 = 900.

\end{WrappedVerbatim}

\noindent\textbf{Reflective Debugging (The ``Aha!'' Moment):}
\begin{WrappedVerbatim}
Thus total assignments = 1+1+900 = 902. Wait this matches solution 2 (13806??). Actually solution 2 gave 902 after inclusion-exclusion? It gave 902 as N_both, but then added others to get 13806. But we argued that assignments where there are empty rows or empty columns are impossible because maximality forces no empties. But earlier we argued that any chip presence forces all rows and columns coloured. Let's verify: Could we have a configuration where some rows are empty, but all columns are coloured (i.e., have chips), and still maximal? Let's test.
\end{WrappedVerbatim}

\begin{WrappedVerbatim}
Suppose a column j is coloured black (has chips). If there is an empty row i (no chips), then intersection (i,j) is empty. Could we add a black chip there? Row i currently empty, no colour restriction; adding black chip would make row i coloured black, which would be consistent because column j is black. So maximality would be violated. Therefore empty rows cannot coexist with any coloured column. If there are empty rows, then all columns must be empty. But then no chips at all, which is not maximal because we could add a chip anywhere. So empty rows not allowed.

Similarly empty columns not allowed.

Thus indeed both rows and columns must all be coloured. So the set of colours used among rows must be same as among columns, and both sets nonempty (since at least one chip). Therefore total assignments = 902.

Wait: but earlier we considered assignment where rows white, columns white yields full grid white. That's okay. What about rows white, columns white except some columns black? That would make black columns have no matching white rows, so they'd be empty (since they cannot have chips). But columns must be coloured (contain at least one chip) because we argued no empty columns. So black columns not allowed unless there is at least one white row (there is) but row colour mismatch, so they cannot have chips. So black columns would be empty, contradiction. Therefore columns cannot have a colour that doesn't appear among rows. So the colour set must be same.

Thus count 902 is correct.

But we must also ensure chip supply: For each colour, number of chips used = (#rows of that colour)*(#columns of that colour). Since each factor <=5, product <=25, okay.

Thus answer 902.

Let's double-check: Are there any configurations where a row is coloured but column set includes both colours, but some columns of same colour as row may be zero? For each row colour, there must be at least one column of same colour, otherwise that row would have zero chips. Since both colours appear among columns, each row colour will have at least one column of same colour (maybe not, if columns have both colours, each colour appears at least once, so yes). So condition satisfied.

Thus total assignments = 902.

Edge case: Could we have rows all white, columns all white? That's counted as 1. Similarly all black.

Thus answer 902.
\end{WrappedVerbatim}
\end{small}

\noindent\textbf{Reasoning path (Round 2 Reviewer 3; answer = 902, correct; recovery).}
\begin{small}
\begin{WrappedVerbatim}
We need to evaluate which solution is correct. Problem: 25 white chips indistinguishable, 25 black chips. Place some chips in 5x5 grid, each cell at most one chip, all chips in same row and same column have same colour, and configuration is maximal: any additional chip would violate one of the previous two conditions.

Interpretation: For any row that contains at least one chip, all chips in that row must be same colour. Similarly for any column. Also, the colour of a chip must be consistent with both its row and column constraints. So a chip at intersection of a row of colour X and column of colour Y must have X=Y, else impossible. Maximal means you cannot add another chip anywhere without breaking condition (either exceed one per cell, or row/col colour uniformity). So any empty cell that could be filled respecting colour uniformity would contradict maximality, so must be impossible to fill.

We need count of maximal placements.

Let's reason.

Let rows be labeled 1-5, columns 1-5.

Define for each row i: either empty (no chips) or assigned colour W or B. Similarly each column j.

If a row i is assigned colour C, then any column j that also has colour C must have chip at (i,j) (otherwise could add chip of colour C). Conversely, if row i colour C and column j colour D != C, then (i,j) must be empty (cannot place chip because would conflict). Also if column j is empty, then any row with colour cannot have chip there because column empty would mean no chip in column; but can we have row coloured and column empty? Suppose row i is white, column j is empty (no chips in column). Then cell (i,j) is empty. Could we add a white chip there? Row i already white, column j empty would then get a white chip, making column j have a chip, thus column j would become white, which would be allowed? But maximality says you cannot add any chip without violating condition. If column j empty, adding a chip would give column j a colour equal to row's colour, which would be consistent (since column would then have all chips white). So to be maximal, we cannot have a row coloured and a column empty, because then you could add chip at intersection. Therefore if any row is coloured, all columns must be coloured (non-empty). Similarly, if any column is coloured, all rows must be coloured. So either all rows and columns are coloured, or all rows empty and all columns empty? But empty grid is not maximal because you could add a chip anywhere. So must have all rows coloured and all columns coloured. Therefore each row and column is assigned either W or B (no empty). This matches consensus of solutions 1 and 3.

Now we need to ensure that chip supply of 25 each is not exceeded. Number of white chips = (#white rows)*(#white columns). Max possible is 5*5=25, okay. Similarly black chips = (#black rows)*(#black columns). Since total cells is 25, chips of both colours sum to at most 25 (actually exactly number of cells occupied = sum of those products). Could exceed 25 for each colour individually? No, each product <=25. So supply fine.

Thus each assignment of colours to rows and columns yields a unique maximal placement: fill all cells where row colour = column colour. This is maximal because any empty cell has mismatched colours, adding chip would violate uniformity (row would now have two colours). So count = 2^5 choices for rows * 2^5 for columns = 2^10 = 1024.

Check if any assignments produce same placement? No, because placement determines row colour set? Actually if a row has no chips (i.e., all its intersections are mismatched), could that happen? Suppose a row is white, but all columns are black. Then row white, columns black, then no cell matches, so row has no chips, violating condition that row is coloured but empty. But earlier we argued that if row coloured and columns all opposite colour, then row would be empty, but maximality? Let's test: Row white, all columns black. Then any cell (row, column) would have mismatched colours, thus empty. Could we add a white chip at (row, column) ? Column is black (has chips maybe). Column black already has chips (since columns are black, they have chips in rows that are black). Adding white chip to black column would give column two colours, violation. So cannot add chip. So configuration could be maximal even though a coloured row has zero chips? Wait row white has zero chips because no column white. But row is coloured white, yet no chip in that row. Is that allowed? The condition "all chips in the same row have same colour" does not require that row with chips must have at least one chip. So a row could be assigned a colour but be empty, as long as no chip violates. However earlier we argued that a coloured row with empty column leads to possibility to add chip. Here columns are not empty; they have chips (black). Adding white chip would conflict column colour. So maximality holds. So our earlier deduction that empty rows cannot coexist with nonempty columns may be too strong. Need to revisit.
\end{WrappedVerbatim}
\end{small}

\noindent\textbf{Reasoning path (Round 3 Reviewer 1; answer = 902, correct).}
\begin{small}
\begin{WrappedVerbatim}
We need to analyze these proposed solutions and determine the correct answer. Problem: 25 white chips, 25 black chips, indistinguishable within each colour. 5x5 grid, at most one chip per cell. Condition: all chips in same row and all chips in same column have same colour. Also maximal: any additional chip would violate one of the previous two conditions.

We need count of ways to place some chips satisfying those constraints, using up to 25 of each colour (since we have that many). Actually we have exactly 25 of each available, but we can use any number <=25.

We need to find number of maximal placements.

The solutions differ: some say 1024 (2^10) counting all colour assignments to rows and columns, allowing possibly all rows same colour, all columns same colour, etc. Others argue that both colours must appear (except all-white or all-black) because otherwise you could add a chip? Let's examine.

We need to ensure maximality: cannot add any chip without violating uniform colour condition (row/col uniform) or at most one per cell.

If all rows and columns are white, then the board is fully filled with white chips (since every intersection of same colour). That's 25 white chips, uses all white chips, leaves black chips unused. Is this maximal? Could we add a black chip somewhere? No, because any cell already occupied. Could we add a white chip? No empty cells. So maximal yes. So all-white configuration is allowed.

Similarly all-black.

Now consider a configuration where rows and columns are coloured arbitrarily, but each row and column must contain at least one chip (as argued). However, is it possible to have a configuration where some rows are white, some black, and columns similarly, and still be maximal? Let's test.

Suppose we have rows: first row white, second row black. Columns: first column white, second column black. According to rule, chips placed at intersections where row colour equals column colour: (white row, white column) and (black row, black column). So we place chips at (row1,col1) white, (row2,col2) black, etc. This yields a pattern where some cells are empty because colours differ. Is this configuration maximal? Consider an empty cell where row colour differs from column colour, e.g., row1 white, col2 black. Could we add a chip there? If we add a white chip, row1 would have both white and black (since column2 is black column, but column colour is black, but column would then have both colours because that column already has a black chip at (row2,col2). Adding a white chip would make column2 have both colours, violating condition. Adding a black chip would make row1 have both colours. So cannot add. So indeed maximal.

Thus any assignment of colours to rows and columns (each row coloured either white or black, each column coloured either white or black) yields a maximal configuration, as long as we fill all matching cells.

But is there any restriction that a colour must appear among rows and columns? Suppose we have rows all white, columns mix white and black. Then some columns black, rows white. At intersection of white row and black column, colours differ, no chip placed. However, column black would have no chip because any chip in that column would require a row black to match, but there are none. Therefore that column would be empty, violating earlier claim that every column must contain at least one chip. But is that required? Let's examine condition: maximality: if there is an empty column (no chips), but there is a white row (nonempty), then the intersection cell (white row, black column) is empty, could we place a white chip there? Row is white, column is black (currently empty). Placing a white chip would make column have a white chip, but column would then have a chip of colour white while column's assigned colour (if any) is black. However column didn't have any chip, so its colour is not defined yet. The condition "all chips in the same column have the same colour" does not require a column to have a colour if empty. So we could assign the column's colour to white by placing a white chip; that would not violate anything. So the configuration with all rows white, some columns black (empty) is not maximal because we could add a white chip in those empty columns (at any row, but row is white). That would not violate condition because column would then have only white chips. So maximality forces that there cannot be an empty column while there is a nonempty row. Similarly cannot have empty row while nonempty column. Thus either both rows and columns are all empty (no chips) — but that's not maximal because you could add a chip anywhere — or all rows and columns are nonempty. So each row and column must have at least one chip. That implies each row colour must appear among columns (so there is at least one column of same colour) and each column colour must appear among rows. Hence both colours may not be required; we could have all rows white and all columns white: then each column nonempty (since all rows white, all columns have white chips). That's fine. Could we have all rows white but columns mix? Then some columns black would be empty (since no black rows). That violates condition because there is a nonempty row (white) and empty column (black). So not maximal. So the only assignments allowed are those where the set of colours used among rows equals set used among columns, and each colour used appears at least once in both rows and columns. Thus either only white appears (all rows white, all columns white) or only black appears (all black), or both colours appear with at least one row and one column of each colour.

Thus number of assignments: all-white 1, all-black 1, both colours: number of colourings of rows with both colours present = 2^5 - 2 = 30. Similarly columns 30. Total 30*30 = 900. Sum = 902.

Thus solution 902 is correct.

Now need to verify that supply of chips (max 25 each) is never exceeded. For both colours case, #white chips = (#white rows)*(#white columns) <= 4*4? Actually max white rows 5, white columns 5, product up to 25, okay. Same for black.

Thus answer 902.

Now produce JSON with steps and answer.
\end{WrappedVerbatim}
\end{small}

\noindent\textbf{Reasoning path (Round 3 Reviewer 2; answer = 902, correct).}
\begin{small}
\begin{WrappedVerbatim}
We need determine correct answer. Problem: 5x5 grid, 25 white chips indistinguishable, 25 black chips indistinguishable. Place some chips (could be zero? but condition "any additional chip placed ... would violate" suggests maximal placement). Conditions: each cell at most one chip, all chips in same row and same column have same colour (i.e., each row that contains any chips must be monochromatic, and each column that contains any chips must be monochromatic). Maximality: cannot add another chip without breaking previous conditions.

We need count number of maximal placements.

We have multiple proposed solutions: two give 1024 (assuming all rows and columns colored, no restriction on having both colours) and three give 902 (excluding all-white and all-black? Actually they include those as 1 each, but also require both colours present when both colours appear, they count 30*30=900, total 902). The difference is whether the all-white/all-black configurations are allowed under maximality. Let's analyze.

If all rows and columns are white, then every cell is intersection of white row and white column, so all 25 cells would be occupied by white chips. That's a placement using 25 white chips, 0 black chips. Is it maximal? Could we add a black chip somewhere? No, because each row already has white chips, adding a black chip would break row uniformity. Could we add another white chip? No cells empty. So maximal. Similarly all black.

But does the condition "all chips in the same row and all chips in the same column have the same colour" allow a row to be empty? Yes, row with zero chips trivially satisfies. The condition is about rows that have chips; they must be same colour. So empty rows are allowed.

Now maximality condition: "any additional chip placed on the grid would violate one or more of the previous two conditions." If we have all-white board, we cannot add any chip because all cells occupied. So it's maximal. So all-white configuration should be counted.

But the solutions that got 902 argued that if any row has chips then all columns must have chips, leading to conclusion that no empty rows/columns. However they also argued that the set of colours present among rows equals set among columns, and they required that set be non-empty (i.e., at least one chip). They considered case where both colours appear, they need at least one row of each colour, etc. However they also counted the all-white/all-black cases separately (1 each). So they included them. So 902 includes them.

Why then 1024 excludes something? The 1024 solution counted all 2^5 * 2^5 assignments, including those where rows are all white and columns all black? Wait that would be inconsistent because row white column black leads to empty board? Let's see: If rows are white, columns black, then any cell would have mismatched colours, so no chip can be placed (since chip colour must match both row and column). That would be empty board. Is empty board maximal? Condition: any additional chip placed would violate. If board empty, could we place a chip? Choose any cell: row currently empty, column empty, placing a chip would assign colour to that row and column. Would that violate uniformity? Row would have only that chip, column only that chip, so uniformity holds. So we could add a chip, thus empty board is NOT maximal. So assignments where row colour set != column colour set produce empty board (or partially empty with no intersections). For maximality we need that for each row colour there is at least one column of same colour, otherwise that row would be empty, violating earlier claim that no empty rows allowed. But earlier claim that no empty rows allowed was derived from maximality: if a row is empty but there is a column with chips, you could add. But if all rows are empty and all columns empty (i.e., no chips at all), then you could add a chip, so not maximal. So indeed any maximal configuration must have at least one chip, and each row and column must have at least one chip (as argued). However is it possible to have a row empty while all columns also empty? That's the empty board case, not maximal.

Now consider assignments where rows all white, columns all white: leads to full board, fine. rows all white, columns mixture? Suppose rows all white, columns have both white and black. Then any white column intersect white row => chip placed (white). Black columns intersect white rows mismatched, no chip. So some columns (black) would be empty. But those columns are non-empty? They have no chips. Are they considered empty columns? Yes. Then we have empty columns while rows have chips (white rows have chips). This violates maximality because there is a row with chips (white) and an empty column (black). Intersection cell (row, black column) is empty, but could we add a chip? Row is white, column is black: to place a chip, the chip colour would have to be both white and black impossible. So we cannot add a chip there without violating uniformity. However the earlier argument that any empty column with a non-empty row leads to possibility of adding a chip of row's colour assumes column colour is same as row's colour? Wait they argued you could add chip of row's colour at intersection, but column would then have a chip of row's colour, violating column uniformity if column previously had no chips (i.e., empty). Actually column is empty, so adding a chip of row's colour would make column have that colour, which is okay because column was empty. So column would become non-empty with that colour, uniform. So it's allowed. But does column have a predetermined colour? In model where each column has a colour assigned, if column is empty originally, its colour is undefined. The earlier solution that assigns colours to all rows and columns even empty ones forced each column to have a colour. But maximality does not require preassigned colour; empty column can adopt colour of first chip placed. So the argument that empty column with non-empty row leads to contradiction only holds if we assume column already has a colour that differs. But if column is empty, we could still add chip, making column non-empty, not violating anything. Thus maximality does NOT prohibit empty columns when rows have chips, because you could add chip. Wait but then you'd be able to add chip, contradicting maximality. So maximal configuration cannot have an empty column while there is a non-empty row, because you could add a chip at intersection (assign column colour equal to row's colour). That would be allowed and would not violate uniformity. So indeed maximality forces no empty columns if any row has chips. Similarly no empty rows if any column has chips.

Thus any maximal configuration must have either all rows empty (i.e., no chips) which is not maximal, or all rows non-empty and all columns non-empty. So each row and column must contain at least one chip. Therefore each row and column must have a colour (since they have chips). So the colour assignment model must have each row and column coloured. Moreover, the colour of a row must appear among columns, otherwise that row would have no matching column to place chips, leading to row empty, contradiction. So the set of colours used among rows equals set among columns, and must be non-empty. Could be just white, just black, or both.

Now consider the case where rows all white, columns all white: fine. rows all white, columns all white -> full board white. rows all white, columns all white is allowed. What about rows all white, columns all black? Then rows have chips? For row white, need a column white to intersect. But there are none, so rows would be empty, contradicting each row must have at least one chip. So this assignment invalid. So to have all rows white, columns must include at least one white column. Since columns all black fails. So assignments where rows colour set != columns colour set are invalid because some rows/columns would be empty. Therefore the count should be: number of colourings of rows and columns using colours from {W,B} such that the set of colours used in rows equals set used in columns, and both sets non-empty. That yields: (all white) 1, (all black) 1, (both colours) count = (2^5 -2) * (2^5 -2) = 30*30 = 900. Total 902. This matches the 902 solutions.

Now why 1024 solutions included assignments where rows white, columns black etc., leading to empty board, but they claimed maximality forces every cell whose row colour equals column colour be occupied, and they didn't require each row to have at least one chip. They argued that maximality also forces that every row and column must contain at least one chip, but they gave proof: "If any column were non‑empty while a row were empty, the cell at their intersection could be filled with a chip of the column’s colour, contradicting maximality. The same argument holds for an empty column with a non‑empty row. Hence in any maximal configuration every row and every column must contain at least one chip; i.e., all rows and columns are coloured (no “empty” rows or columns)." This proof is correct. But they then considered assignments where rows white, columns black: then rows are coloured white, columns coloured black, but each row has chips? Wait row white, column black: there is no cell with matching colours, so rows would have zero chips, contradicting earlier conclusion that each row must have at least one chip. So such assignments are invalid. However they counted them anyway because they didn't enforce that each row must have at least one matching column. They assumed that if rows and columns are coloured, then maximality forces each matching cell occupied, but they didn't require existence of any matching cell. So they overcounted.

Thus correct answer is 902.

Now produce steps summarizing consensus and divergent analysis, final answer 902.
\end{WrappedVerbatim}
\end{small}

\noindent\textbf{Reasoning path (Round 3 Reviewer 3; answer = 902, correct).}
\begin{small}
\begin{WrappedVerbatim}
We need to evaluate these proposed solutions and produce correct answer.

Problem: 25 white indistinguishable chips, 25 black chips, place some in 5x5 grid, at most one per cell, all chips in same row and same column have same colour, any additional chip would violate condition (maximal). Count number of ways.

We need to consider configurations satisfying:

- For each row that has at least one chip, all chips in that row are same colour.
- For each column that has at least one chip, all chips in that column are same colour.
- Additionally, the colour of a chip must be consistent with both its row and column: thus if a chip is placed at intersection of row i and column j, the row colour must equal column colour, otherwise impossible.

Thus configuration can be described by assigning each row a colour (or possibly empty row) and each column a colour (or possibly empty column) such that any occupied cell is at intersection of same-colour row and column, and all such intersections must be occupied for maximality (otherwise could add chip). Also we must ensure that there is no empty row with non-empty column (or vice versa) because then intersection would be empty but could be filled.

Let's reason: Suppose we have some rows empty (no chips). If there exists any column with at least one chip, then the intersection of that empty row and that column is empty but could add chip of column colour (since row currently empty, adding chip would set the row colour to that colour, still uniform). That would not violate row uniformity because row would then have only that colour. So maximality fails. Therefore if any row is empty, all columns must be empty. Symmetrically, if any column empty, all rows empty. So either all rows and columns are nonempty, or all rows and columns are empty (no chips at all). But is the empty configuration allowed? Condition: "any additional chip placed on the grid would violate one or more of the previous two conditions." If we have no chips, can we add a chip? Row and column would each have just that chip, uniform. So adding a chip would not violate; thus empty configuration is not maximal. So we must have all rows and columns nonempty. Thus each row and column must have at least one chip, meaning each row assigned a colour, each column assigned a colour.

Now we must also ensure that the set of colours used among rows equals set among columns, otherwise there would be a colour present only in rows but not in any column, causing those rows to have chips? Let's see: Suppose some row is white, but there is no white column. Then any cell in that row would intersect a column of black colour, mismatch, so no chip could be placed in that row (since chip requires row colour = column colour). But row must have at least one chip (nonempty). Contradiction. Therefore each colour that appears among rows must also appear among columns, and vice versa. So the set of colours used by rows equals that used by columns, and must be nonempty.

Thus possible colour sets: {white} only, {black} only, or both {white, black}. For each case, we need to count assignments of colours to rows and columns such that each colour appears at least once among rows and columns (if both colours present). For single-colour case, all rows and columns are that colour: 1 assignment each for white, 1 for black.

For both colours present: Number of ways to colour 5 rows with both colours present = total 2^5 minus the two monochrome assignments = 32-2=30. Similarly columns: 30. All such pairs produce a configuration? Need to verify that maximality holds: For each pair of colourings, all cells where row colour = column colour are filled. Are there any empty cells where row colour = column colour? No by rule we fill them all. Are there any empty cells where row colour != column colour? Those cannot be filled because would cause mismatched colours. So configuration is maximal. Chip counts: white chips = (#white rows)*(#white columns) <= 5*5=25 okay. Black similarly.

Thus total configurations = 1+1+30*30 = 902.

The other solutions with 1024 ignored the requirement that both colours must appear if any row of that colour exists, they allowed assignments where rows are all white but some columns black (or vice versa).
But those would produce empty rows/columns? Let's test: Suppose rows all white, columns some black.
Then any black column intersect white rows gives mismatch, so those cells cannot have chips.
Black column would be empty (no chips) which violates requirement that each column must have at least one chip (since we argued all columns must be nonempty).
Indeed the earlier argument: if a column is empty but there is a nonempty row, you could add chip at intersection.
So such assignments are invalid. So 1024 overcounts those invalid configurations. Therefore correct answer is 902.

Now produce steps summarizing consensus and divergent analysis.

Consensus steps: reasoning that each row/column must be nonempty, colour assignments, chip placement at matching colours, chip counts within limits. Divergent: whether we need to exclude assignments where colour sets differ; correct is to exclude, leading to 902.

Now produce JSON with steps and answer.
\end{WrappedVerbatim}
\end{small}

\noindent\textbf{Reasoning path (Round 3 Reviewer 4; answer = 902, correct).}
\begin{small}
\begin{WrappedVerbatim}
We need to determine correct answer. The problem: 25 white chips, 25 black chips indistinguishable, place some in 5x5 grid, each cell at most one chip, all chips in same row and same column have same colour, maximal (cannot add any chip without violating). Need count of ways.

We have multiple proposals: 1024 (solutions 1 and 4) and 902 (solutions 2 and 3). Need to analyze divergent point: whether empty rows/columns allowed.
The first solution claims every row and column must contain at least one chip (i.e., no empty rows/columns).
Actually they argue that if any column non-empty while a row empty, you could place chip at intersection, violating maximality.
But that assumes you can place chip of column's colour in that row, but row is empty, so row would then have a chip of that colour; that wouldn't conflict with row uniformity (since only one chip).
However, condition also says "all chips in the same row have the same colour". An empty row has no chips, so adding one chip of column's colour would be okay.
So maximality indeed forces no empty row while there exists any non-empty column. Therefore either all rows empty (no chips at all) or all rows non-empty.
But we also need columns non-empty similarly. So either no chips at all (empty board) or all rows and columns non-empty.

But is empty board considered maximal? The condition "any additional chip placed on the grid would violate one or more of the previous two conditions." If board empty, can we add a chip? Adding a single chip would satisfy both conditions (row uniform, column uniform) because only one chip in its row/col. So empty board is NOT maximal. So we must have at least one chip, thus all rows and all columns must be non-empty. So each row and column has at least one chip.

Thus each row and column must be assigned a colour (since they have chips, they have a colour). So each row colour, each column colour. The placement then must be all cells where row colour equals column colour (since any such cell must be occupied, otherwise we could add a chip there without breaking uniformity). So each colour assignment yields a unique placement.

Now counting colour assignments: 2 choices per row, 2 per column: 2^10 = 1024. But we must ensure that the resulting placement does not exceed chip supplies: number of white chips = (#white rows)*(#white columns) <=25; similarly black chips = (#black rows)*(#black columns) <=25. Since each count max 5*5=25, always okay. So all 1024 assignments are valid.

But the alternative solutions subtract the all-white and all-black assignments? They consider that set of colours present among rows = set among columns must be non-empty, but also they think both colours must appear?
They argue that if all rows and columns are white, then that's a maximal placement? Let's examine: If all rows and columns are white, then all 25 cells are white chips, using all white chips.
That's a valid configuration. Is it maximal? Could we add a black chip somewhere? Adding a black chip would conflict because its row already has white chips, so row would have two colours, violating condition.
So cannot add any chip of any colour. So it's maximal. So the all-white and all-black configurations are valid.
The 902 answer excluded them? Actually they counted them as 1+1 and added 30*30 for both colours present, giving 902. That's 2 + 900 = 902.
The 1024 answer counts all 2^10 = 1024, which includes the all-white and all-black cases, as well as other assignments where one colour may be absent?
Wait if a colour absent among rows, then all rows same colour, say all white, but columns could have both colours?
But earlier we argued that set of colours among rows must equal set among columns, otherwise some column colour would not appear among rows leading to empty row?
Let's test: Suppose rows all white, but columns some black, some white. Then consider a black column; its intersection with any white row would be a cell where row white, column black, colours mismatch, so that cell must be empty (cannot place chip).
But column black would have no chip because any chip in that column must match row colour, which are white, not black. So column black would be empty, contradicting requirement that each column non-empty.
Indeed earlier we argued each column must be non-empty, so a column coloured black must have at least one chip, which would require a row of black. Since rows all white, impossible.
Thus columns cannot have a colour not present among rows. So the sets must be equal. So if rows all white, columns must all be white as well.
So the only assignments where a colour is absent are the all-white or all-black configurations. Those are counted in 2^10.
So the 902 answer excluded those? No they included them as +1+1. So why difference? Because they counted assignments where both colours appear: rows have both colours (2^5-2=30) and columns both colours (30), giving 900. Add 2 = 902.
But 2^10 = 1024 includes also assignments where rows have both colours but columns maybe all white? That would be impossible due to non-empty condition.
But 2^10 counts all 1024 assignments regardless of that constraint. Let's see if there are assignments where rows have both colours but columns are all white.
That would be counted in 2^10 but invalid because columns all white but rows have black rows -> black rows intersect white columns mismatched, causing black rows empty?
Actually black rows would have no chip because any column is white, so black rows cannot have chip, violating non-empty column condition?
Wait columns all white means each column colour white, each column must have at least one chip; that chip would be white, placed at intersection with any white row (since row must be white to match).
Black rows would have no matching columns, thus black rows would be empty, violating condition that each row non-empty. So such assignment invalid.
Thus 2^10 overcounts assignments where the sets of colours differ. How many such assignments exist?
\end{WrappedVerbatim}

\noindent\textbf{Reflective Debugging (The ``Aha!'' Moment):}
\begin{WrappedVerbatim}
Count assignments where rows and columns colour sets are not equal, i.e., one set is subset of other but not equal, because both must be non-empty.
Possibilities: rows all white, columns mix both colours (invalid). rows mix both, columns all white (invalid). Similarly for black.
So we need to count those invalid assignments and subtract from 1024 to get correct count.

Let's compute: Number of assignments where rows all white (1 way) and columns any (2^5 =32) -> total 32. Similarly rows all black (1*32) =32. Similarly columns all white (32) and rows any (32) = but we already counted the case rows all white columns all white twice. Need careful inclusion-exclusion.

Let's define sets:
A = assignments where rows are monochrome (all same colour). There are 2 choices for colour * 1 way for rows =2. For each such, columns can be any of 2^5=32. So |A| = 2*32 =64.

B = assignments where columns are monochrome. Similarly |B| = 2*32 =64.

Intersection $A \\cap B$ = assignments where both rows and columns monochrome. There are 2 possibilities (all white or all black). So $|A \\cap B|=2$.

Now assignments where row set != column set are those in $(A \\cup B) \\setminus (A \\cap B)$ plus also assignments where both are non-monochrome but still sets differ. Could that happen? If rows have both colours, then set is $\\{W,B\\}$. For columns to differ, they'd have to be monochrome (since the only other set is a single colour). So that case is covered by $B$ but not $A$. Similarly columns both colours, rows monochrome covered by $A$. So indeed any unequal sets involve at least one side monochrome. Thus the invalid assignments are $(A \\cup B)$ minus the cases where both monochrome with the same colour. Those are in $A \\cap B$ and should be kept. So
\\[
\\text{invalid} = (A \\cup B) \\setminus (A \\cap B) = |A| + |B| - 2|A \\cap B| = 64 + 64 - 4 = 124.
\\]

Thus valid assignments = 2^10 - 124 = 1024 -124 = 900. Wait that's 900, not 902. But earlier we had 902 valid. Let's double-check.

But we also need to consider assignments where rows all white and columns all black (or vice versa). Those are in A \\cap B? Actually rows all white, columns all black: rows monochrome white, columns monochrome black. That's in A (rows monochrome) and also in B (columns monochrome) but not same colour, so not in intersection of same colour. Intersection defined as both monochrome (any colour) includes this case? Intersection of A and B means both rows monochrome and columns monochrome, regardless of colour. So includes both-colour mismatched cases. But those configurations are invalid because rows white, columns black: then there are no cells where colours match, so no chip placed, violating non-empty condition (since rows and columns would be empty). Actually each row white but no matching column white, so rows have no chips, contradicting requirement that each row non-empty. So these are invalid and should be removed. Our earlier subtraction counted them as invalid (since they are in A \\cap B but not the two all-white/all-black). So we need to keep only the two all-white/all-black. So we should subtract all of A \\cup B except those two.

Let's compute number of assignments that are valid: those where row set = column set (both non-empty). Cases:

- Both sets = {W}: all rows white, all columns white => 1 assignment.
- Both sets = {B}: all rows black, all columns black => 1 assignment.
- Both sets = {W,B}: rows have both colours, columns have both colours. Need rows not all white nor all black (i.e., 2^5 -2 =30). Similarly columns 30. So 30*30 =900 assignments.

Total = 902. That matches solution 2/3.

Now why did we earlier think rows must be non-empty? That's satisfied. The invalid assignments counted in 1024 are those where sets differ: either rows monochrome but columns not same monochrome, etc. Let's count number of such invalid assignments: total 1024 - 902 =122. Let's compute again: A \\cup B includes assignments where at least one side monochrome. Count of assignments where rows monochrome (any colour) = 2*2^5 =64 as before. Similarly columns monochrome =64. Intersection (both monochrome) = 2*2 =4 (since rows colour 2 choices, columns colour 2 choices). Wait earlier we said intersection size 2 (both same colour). But both monochrome includes 4 combos: (rows white, columns white), (rows white, columns black), (rows black, columns white), (rows black, columns black). So |A \\cap B|=4.

Thus |A \\cup B| = 64+64-4 =124. Among these, the two valid all-white and all-black are part of A \\cup B (they are both monochrome). So invalid = 124 -2 =122. Good matches difference.

Thus correct answer is 902.

We must ensure that configurations where rows both colours but columns monochrome are indeed invalid because columns would be non-empty? Let's test rows both colours, columns all white. Columns all white, each column must have at least one chip. Since rows have black rows, those black rows intersect white columns mismatch, so black rows would have no chip, violating row non-empty. So invalid.

Thus final answer 902.

Now produce JSON with steps and answer.
\end{WrappedVerbatim}
\end{small}

\section{MMLU-PRO per-category results (Appendix)}
\label{app:mmlu-category-table}
\noindent Table~\ref{tab:result1} reports MMLU-PRO overall. Below we provide a per-category breakdown in the same format: Sampling vs.\ \textbf{DCR (Single)} and \textbf{DCR (Mix)}.

\begin{table}[!htbp]
\centering
\small
\setlength{\tabcolsep}{5pt}
\renewcommand{\arraystretch}{1.10}
\caption{MMLU-PRO per-category trial accuracy and consistency (\%), part 1 of 3.}
\label{tab:mmlu_cat_exp1_part_1}
\begin{tabular}{llrrrr@{\hspace{6pt}}rrr}
\toprule
\textbf{Category} & \textbf{Model} & \textbf{N} & \multicolumn{3}{c}{\textbf{Trial acc. (\%)}} & \multicolumn{3}{c}{\textbf{Consistency (\%)}} \\
 &  &  & \textbf{Samp.} & \textbf{DCR(S)} & \textbf{DCR(M)} & \textbf{Samp.} & \textbf{DCR(S)} & \textbf{DCR(M)} \\
\cmidrule(lr){4-6}\cmidrule(lr){7-9}
\midrule
biology & GPT-OSS & 717 & 89.9 & 90.8 & 91.1 & 90.7 & 90.8 & 91.8 \\
biology & Llama-4 & 717 & 85.4 & 88.4 & 90.2 & 85.6 & 88.3 & 90.5 \\
biology & Llama-3.3 & 717 & 85.2 & 86.7 & 89.8 & 85.8 & 86.6 & 90.0 \\
biology & Granite-4 & 717 & 74.1 & 74.8 & 87.1 & 74.1 & 74.5 & 87.3 \\
business & GPT-OSS & 789 & 84.5 & 86.3 & 85.1 & 85.7 & 86.7 & 87.1 \\
business & Llama-4 & 789 & 56.7 & 81.3 & 84.8 & 56.3 & 81.1 & 85.7 \\
business & Llama-3.3 & 789 & 72.5 & 78.1 & 83.1 & 73.1 & 78.0 & 83.4 \\
business & Granite-4 & 789 & 32.4 & 36.5 & 74.1 & 31.2 & 36.4 & 75.0 \\
chemistry & GPT-OSS & 1132 & 87.5 & 88.9 & 85.8 & 87.8 & 88.8 & 87.2 \\
chemistry & Llama-4 & 1132 & 54.8 & 80.8 & 85.3 & 55.0 & 80.9 & 87.2 \\
chemistry & Llama-3.3 & 1132 & 68.2 & 76.8 & 81.2 & 70.0 & 76.7 & 81.9 \\
chemistry & Granite-4 & 1132 & 31.8 & 35.0 & 72.2 & 30.1 & 34.6 & 72.6 \\
computer science & GPT-OSS & 410 & 84.8 & 86.2 & 81.3 & 84.9 & 86.6 & 83.2 \\
computer science & Llama-4 & 410 & 67.0 & 77.5 & 81.9 & 67.3 & 77.6 & 83.2 \\
computer science & Llama-3.3 & 410 & 72.5 & 75.4 & 73.4 & 73.2 & 75.4 & 72.4 \\
computer science & Granite-4 & 410 & 53.0 & 54.4 & 75.2 & 52.2 & 53.7 & 75.1 \\
economics & GPT-OSS & 844 & 83.5 & 85.6 & 85.3 & 84.1 & 85.7 & 85.8 \\
economics & Llama-4 & 844 & 78.6 & 84.5 & 85.3 & 78.8 & 84.5 & 86.0 \\
economics & Llama-3.3 & 844 & 78.6 & 80.4 & 84.4 & 79.0 & 80.5 & 84.4 \\
economics & Granite-4 & 844 & 61.4 & 62.7 & 80.1 & 61.3 & 62.6 & 80.9 \\
engineering & GPT-OSS & 969 & 68.8 & 73.7 & 75.6 & 68.9 & 73.0 & 76.6 \\
engineering & Llama-4 & 969 & 57.1 & 69.6 & 74.6 & 57.2 & 69.6 & 75.3 \\
engineering & Llama-3.3 & 969 & 53.2 & 59.9 & 71.1 & 54.7 & 60.1 & 71.0 \\
engineering & Granite-4 & 969 & 38.4 & 41.0 & 64.0 & 37.6 & 40.5 & 64.0 \\
health & GPT-OSS & 818 & 76.1 & 77.4 & 78.2 & 75.9 & 77.6 & 78.4 \\
health & Llama-4 & 818 & 76.3 & 77.7 & 78.2 & 76.4 & 77.6 & 78.2 \\
health & Llama-3.3 & 818 & 72.7 & 75.6 & 78.4 & 73.1 & 75.8 & 79.0 \\
health & Granite-4 & 818 & 57.0 & 58.4 & 74.6 & 57.6 & 58.6 & 74.3 \\
history & GPT-OSS & 381 & 66.2 & 68.2 & 69.5 & 65.9 & 67.7 & 69.5 \\
history & Llama-4 & 381 & 69.3 & 69.3 & 68.1 & 69.0 & 69.3 & 68.0 \\
history & Llama-3.3 & 381 & 63.8 & 64.5 & 68.8 & 64.0 & 64.8 & 69.0 \\
history & Granite-4 & 381 & 50.0 & 51.1 & 65.9 & 48.6 & 50.9 & 66.7 \\
law & GPT-OSS & 1101 & 54.4 & 59.2 & 58.9 & 55.6 & 59.6 & 59.3 \\
law & Llama-4 & 1101 & 56.4 & 57.0 & 59.1 & 56.5 & 57.1 & 59.2 \\
law & Llama-3.3 & 1101 & 51.0 & 53.0 & 57.0 & 50.7 & 52.7 & 57.1 \\
law & Granite-4 & 1101 & 32.1 & 32.5 & 52.5 & 32.0 & 32.6 & 53.2 \\
math & GPT-OSS & 1351 & 93.5 & 94.5 & 87.1 & 94.5 & 94.5 & 90.8 \\
math & Llama-4 & 1351 & 56.3 & 86.2 & 88.5 & 56.4 & 86.2 & 90.8 \\
math & Llama-3.3 & 1351 & 73.8 & 82.2 & 83.1 & 76.5 & 82.1 & 83.1 \\
math & Granite-4 & 1351 & 35.2 & 40.2 & 78.7 & 34.7 & 40.1 & 80.0 \\
other & GPT-OSS & 924 & 73.7 & 75.3 & 76.1 & 74.8 & 75.2 & 76.8 \\
other & Llama-4 & 924 & 64.7 & 73.7 & 76.9 & 64.3 & 73.7 & 77.4 \\
other & Llama-3.3 & 924 & 69.2 & 72.2 & 74.4 & 69.3 & 72.1 & 74.2 \\
other & Granite-4 & 924 & 48.1 & 50.0 & 68.5 & 47.6 & 50.2 & 69.0 \\
philosophy & GPT-OSS & 499 & 68.1 & 70.8 & 72.6 & 68.9 & 70.7 & 72.5 \\
philosophy & Llama-4 & 499 & 68.2 & 72.2 & 72.7 & 67.9 & 72.3 & 73.0 \\
philosophy & Llama-3.3 & 499 & 62.2 & 65.2 & 71.0 & 62.7 & 65.3 & 70.7 \\
philosophy & Granite-4 & 499 & 51.0 & 52.4 & 63.0 & 50.9 & 52.7 & 62.9 \\
\bottomrule
\end{tabular}
\end{table}

\begin{table}[!htbp]
\centering
\small
\setlength{\tabcolsep}{5pt}
\renewcommand{\arraystretch}{1.10}
\caption{MMLU-PRO per-category trial accuracy and consistency (\%), part 2 of 3.}
\label{tab:mmlu_cat_exp1_part_2}
\begin{tabular}{llrrrr@{\hspace{6pt}}rrr}
\toprule
\textbf{Category} & \textbf{Model} & \textbf{N} & \multicolumn{3}{c}{\textbf{Trial acc. (\%)}} & \multicolumn{3}{c}{\textbf{Consistency (\%)}} \\
 &  &  & \textbf{Samp.} & \textbf{DCR(S)} & \textbf{DCR(M)} & \textbf{Samp.} & \textbf{DCR(S)} & \textbf{DCR(M)} \\
\cmidrule(lr){4-6}\cmidrule(lr){7-9}
\midrule
physics & GPT-OSS & 1299 & 87.4 & 88.9 & 86.8 & 88.1 & 88.6 & 88.2 \\
physics & Llama-4 & 1299 & 58.7 & 80.6 & 85.1 & 58.5 & 80.5 & 87.1 \\
physics & Llama-3.3 & 1299 & 68.6 & 76.3 & 83.3 & 69.4 & 76.5 & 83.3 \\
physics & Granite-4 & 1299 & 36.0 & 38.9 & 73.1 & 34.8 & 38.5 & 73.7 \\
psychology & GPT-OSS & 798 & 79.0 & 80.2 & 80.6 & 79.6 & 80.8 & 81.0 \\
psychology & Llama-4 & 798 & 80.2 & 81.0 & 80.1 & 80.2 & 81.0 & 80.6 \\
psychology & Llama-3.3 & 798 & 77.2 & 78.2 & 80.5 & 77.8 & 78.2 & 80.7 \\
psychology & Granite-4 & 798 & 66.3 & 66.7 & 77.1 & 66.4 & 66.4 & 77.3 \\
\bottomrule
\end{tabular}
\end{table}

\end{document}